\documentclass[a4paper,fleqn]{cas-dc}

\usepackage{diagbox}

\usepackage{framed,multirow}

\usepackage{amssymb}
\usepackage{latexsym}

\usepackage{url}
\usepackage{xcolor}
\usepackage{hyperref}
\usepackage{booktabs}
\usepackage{caption}
\usepackage{stfloats}
\usepackage{subcaption}
\usepackage{amsmath}
\usepackage{graphicx}

\usepackage{hyperref}
\usepackage{bbding}
\usepackage{makecell}
\usepackage[font=footnotesize,labelformat=simple]{subcaption}

\usepackage[authoryear,longnamesfirst]{natbib}

\def\tsc#1{\csdef{#1}{\textsc{\lowercase{#1}}\xspace}}
\tsc{WGM}
\tsc{QE}

\begin{document}
\let\WriteBookmarks\relax
\def\floatpagepagefraction{1}
\def\textpagefraction{.001}

\shorttitle{STS MICCAI 2023 Challenge}    

\shortauthors{Y. Wang et al.}  

\title [mode = title]{STS MICCAI 2023 Challenge: Grand challenge on 2D and 3D semi-supervised tooth segmentation}  

\author[1]{Yaqi Wang}
\cormark[1]
\cortext[1]{Corresponding author}
\ead{wangyaqi@cuz.edu.cn}
\address[1]{College of Media Engineering, Communication University of Zhejiang, Hangzhou, China}

\author[2,3]{Yifan Zhang}
\address[2]{State Key Laboratory of Oral Diseases, National Clinical Research Center for Oral Diseases, West China Hospital of Stomatology, Sichuan University, Chengdu, China}
\address[3]{Lishui University, School of Medicine, Hangzhou Geriatric Stomatology Hospital, Hangzhou Dental Hospital Group, Hangzhou, China}

\author[1,4]{Xiaodiao Chen}
\address[4]{School of Computer Science and Technology, Hangzhou Dianzi University, Hangzhou, China}

\author[5,6]{Shuai Wang}
\address[5]{School of Cyberspace, Hangzhou Dianzi University, Hangzhou, China}
\address[6]{Suzhou Research Institute of Shandong University, Suzhou, China}

\author[7]{Dahong Qian}
\address[7]{School of Biomedical Engineering, Shanghai Jiao Tong University, Shanghai, China}

\author[4]{Fan Ye}

\author[4]{Feng Xu}

\author[8]{Hongyuan Zhang}
\address[8]{School of Biomedical Engineering, Medical School, Shenzhen University, Shenzhen, China}

\author[9]{Qianni Zhang}
\address[9]{School of Electronic Engineering and Computer Science, Queen Mary University of London, London, United Kingdom}

\author[10]{Chengyu Wu}
\address[10]{Department of Mechanical, Electrical and Information Engineering, Shandong University, Weihai, China}

\author[4]{Yunxiang Li}

\author[9]{Weiwei Cui}

\author[1]{Shan Luo}

\author[11]{Chengkai Wang}
\address[11]{School of Management, Hangzhou Dianzi University, Hangzhou, China}

\author[1]{Tianhao Li}

\author[12]{Yi Liu}
\address[12]{Department of Stomatology, Sichuan Provincial People's Hospital, University of Electronic Science and Technology of China, Chengdu, China}

\author[4]{Xiang Feng}

\author[13]{Huiyu Zhou}
\address[13]{School of Computing and Mathematical Sciences, University of Leicester, Leicester, United Kingdom}


\author[14]{Dongyun Liu}
\author[15]{Qixuan Wang}
\author[16]{Zhouhao Lin}
\author[17]{Wei Song}

\author[18]{Yuanlin Li}
\author[19]{Bing Wang}
\author[20]{Chunshi Wang}
\author[21,22]{Qiupu Chen}
\author[23]{Mingqian Li}

\address[14]{Zeta Technology Co., Ltd. , No. 1158, Zhangdong Road, Pudong New Area, Shanghai, China}
\address[15]{China Academy of Information and Communications Technology, Beijing, China}
\address[16]{HangZhou Dianzi University, Xiasha Higher Education Zone, Hangzhou, China}
\address[17]{Southeast University, Sipailou, Xuanwu District, Nanjing, China}

\address[18]{Shanghai Ninth People's Hospital, Shanghai Jiao Tong University School of Medicine,Shanghai Jiao Tong University, Shanghai, China}
\address[19]{School of Computer Science and Technology, Changchun University of Science and Technology, Changchun, China}
\address[20]{School of Artificial Intelligence, Guilin University of Electronic Technology, Guilin, China}
\address[21]{University of Science and Technology of China, Hefei, China}
\address[22]{Hefei Institutes of Physical Science, Chinese Academy of Sciences, No. 350, Shushanhu Road, Hefei, China}
\address[23]{School of Information and Optoelectronic Science and Engineering, South China Normal University, Guangzhou, China}
\begin{abstract}
Computer-aided design (CAD) tools are increasingly popular in modern dental practice, particularly for treatment planning or comprehensive prognosis evaluation.
In particular, the 2D panoramic X-ray image efficiently detects invisible caries, impacted teeth and supernumerary teeth in children,
while the 3D dental cone beam computed tomography (CBCT) is widely used in orthodontics and endodontics due to its low radiation dose.
However,
there is no open-access 2D public dataset for children’s teeth and no open 3D dental CBCT dataset, which limits the development of automatic algorithms for segmenting teeth and analyzing diseases.
The Semi-supervised Teeth Segmentation (STS) Challenge,
a pioneering event in tooth segmentation,
was held as a part of the MICCAI 2023 ToothFairy Workshop on the Alibaba Tianchi platform.
This challenge aims to investigate effective semi-supervised tooth segmentation algorithms to advance the field of dentistry. In this challenge, we provide two modalities including the 2D panoramic X-ray images and the 3D CBCT tooth volumes.
In Task 1, the goal was to segment tooth regions in panoramic X-ray images of both adult and pediatric teeth.
Task 2 involved segmenting tooth sections using CBCT volumes.
Limited labelled images with mostly unlabelled ones were provided in this challenge prompt using semi-supervised algorithms for training.
In the preliminary round, the challenge received registration and result submission by 434 teams, with 64 advancing to the final round.
This paper summarizes the diverse methods employed by the top-ranking teams in the STS MICCAI 2023 Challenge.
\end{abstract}




\begin{keywords}
Tooth Segmentation \sep Semi-supervised Learning \sep CBCT \sep Panoramic X-ray
\end{keywords}

\maketitle

\section{Introduction}
In recent years, an increasing number of studies have demonstrated the close association between oral health and numerous systemic illnesses, such as cardiovascular disease \cite{batty2018oral}, diabetes \cite{sanz2018scientific}, and even Alzheimer's disease \cite{dominy2019porphyromonas}. The prevalence of oral diseases in different age groups has also led to other adverse health effects, including body-image issues, sleeplessness, social isolation, pain, discomfort, fear, anxiety, and functional limitations \cite{Y-1,fleming2018prevalence}. Therefore, oral and dental diseases encompass more than just the facio-maxillary region; they also directly impact an individual's overall health and well-being. The worldwide prevalence of oral disease imposes significant challenges and burdens on health care systems. The World Health Organization reports that 358 million people experience severe periodontal conditions, while two-thirds of the global population suffer from dental caries \cite{Y-1}. Such conditions negatively impact the quality of life and exert substantial stress on healthcare systems. As a result, the efficient prevention and treatment of oral diseases, predominantly dental diseases, has emerged as an imperative subject in the public health sector.


Fortunately, the development of Panoramic X-ray Imaging (PXI) and Cone Beam Computed Tomography (CBCT) has changed the way dentists diagnose oral disease, leading to improved diagnostic accuracy, treatment effectiveness, and patient outcomes. PXI has long been a staple in dental practices, offering a comprehensive overview of a patient's oral cavity \cite{Y-2}. It provides dentists with a broad view that encompasses all teeth, the jaws, and other important structures in a single image. This wide-ranging perspective is invaluable for initial evaluations, enabling the identification of various dental issues such as impacted teeth, bone abnormalities, and the overall alignment of the teeth and jaw. PXI's ability to quickly and efficiently provide a general overview of dental health makes it an essential tool for routine check-ups and preliminary assessments. On the other hand, Cone Beam Computed Tomography (CBCT) brings a different set of advantages, primarily its capacity to deliver three-dimensional images. This depth of detail offers clarity regarding the anatomy of the teeth, bones, and soft tissues, allowing for a more precise diagnosis and treatment planning \cite{Y-3,li2021agmb,li2021gt}. CBCT is particularly beneficial in complex cases where spatial relationships and detailed anatomy need to be understood, such as in implantology, orthodontics, and endodontics \cite{gupta2013cone, shukla2017role, uraba2016ability}. Both PXI and CBCT play pivotal roles in dental diagnostics, where PXI is unmatched in providing a quick, comprehensive overview of the dental and maxillofacial region, making it an indispensable tool for general dental assessments. Meanwhile, CBCT excels in cases where three-dimensional detail is crucial for the diagnosis and treatment planning, offering insights that are not possible with two-dimensional imaging alone. 

Despite the great advances in tooth image technology, the visual analysis of such images requires significant time and the expertise of dental experts. It is not only a time-consuming and tedious task, but it also suffers from being prone to errors and subjectivity. These issues limit the potential of tooth imaging technology in clinical research and practice. In this context, deep learning-based image analysis methods have been developed and demonstrated outstanding capability to perform dental structure segmentation, classification, and identification of several common dental diseases with high accuracy \cite{Y-4}. By automatically extracting and learning complex features from dental images, deep learning models provide reliable and reproducible predictions as evidence in various forms to support experts in diagnosis and treatment. Although deep learning has achieved remarkable results in the field, tooth segmentation in 2D panoramic tooth images and 3D CBCT images remains an open problem due to several challenging aspects.

Among these, a primary challenge is the lack of publicly available datasets based on PXI and CBCT to train and validate algorithms and provide benchmarks. To tackle this challenge, we have collected PXI and CBCT datasets from Hangzhou Dental Hospital and Hangzhou QianTang Dental Hospital. These datasets are organized, processed, and then manually annotated with tooth masks that are suitable for training and evaluation. Based on these datasets, we proposed a tooth image segmentation competition in the 26th International Conference on Medical Image Computing and Computer Assisted Intervention (MICCAI 2023), namely, the Semi-supervised Teeth Segmentation (STS) Challenge, hosted on the Alibaba Cloud Tianchi platform. The STS Challenge entails two subtasks, i.e., tooth segmentation in the 2d Panoramic X-ray Image (2D-PXI) dataset and in the 3d dental Cone Beam Computed Tomography (3D-CBCT) dataset.

Another major disadvantage common to all deep learning based computer vision techniques is the dependence on a large quantity of training data, which is hard to obtain, as well as the heavy load on computing resources due to the data scale. The acquisition of 2D-PXI and 3D-CBCT datasets cost a considerable amount of work for professionals in annotation, which is an expensive and time-consuming endeavor. To alleviate the demand for data annotation in the future, we would like to take advantage of this challenge to stimulate the development of semi-supervised learning strategies \cite{X-10,x-11,X-12,X-13,X-14,X-15,X-16,X-17}, which can train models with a comparatively small amount of labelled data alongside a majority of unlabelled data. Such strategies can provide an effective solution to tasks involving few annotated data, thereby allowing deep learning models to achieve excellent performance with less dependence on training data annotation. Therefore, in comparison with the fully supervised approaches, STS Challenge prioritizes the semi-supervised deep learning approaches, which hold greater potential for application in real clinical research tasks. In conclusion, the STS Challenge offers a platform for participants to explore tooth segmentation algorithms and models based on 2D-PXI and 3D-CBCT datasets using semi-supervised deep learning.

In this paper, Section 2 gives a review on the recent research related to semi-supervised medical image segmentation, tooth panoramic image segmentation, and CBCT image segmentation. Section 3 provides a detailed description of the challenge, including the dataset information and evaluation indicators. Section 4 presents the results submitted by the participating teams. Section 5 provides a statistical analysis and evaluation of the Dice, mIoU, and HD metrics of the top ten teams.
The Discussion Section offers some further insights into the challenge results, and the paper concludes with a final remark in Conclusion. 


\section{Related Works}
\subsection{Semi-supervised Medical Image Segmentation}
The development of deep learning technology brings a new era for various tasks in medical image processing. Specifically, medical image segmentation is one of the main beneficiaries of deep learning. It refers to the allocation of all pixels in the image into a certain category so as to accurately depict the required structures in the original image, such as designated organs or lesions.

Based on the success of the U-Net network in 2015 \cite{X-1}, several enhanced networks, such as UNet++ \cite{X-2} and UNet 3+ \cite{X-3}, have been proposed for image segmentation.
Further, Isensee et al. \cite{X-7} introduced nnU-net, an adaptive neural network architecture that can automatically adjust network structure and training strategies to adapt to different 3D and 2D medical image segmentation tasks.
This approach eliminates the need for manual hyperparameter tuning and network structure modifications and, in many cases, outperforms manually optimized networks. With the emergence of ViT \cite{X-4}, researchers combine the transformer structure with the CNN structure, resulting in MT-UNet \cite{X-5}, which has good segmentation performance in abdominal organ segmentation,
and UNETR \cite{X-6}, which is targeted at brain segmentation.

Although the medical image segmentation task has witnessed great successes with these network structures, their excellent performance is highly dependent on high-quality, large-scale labelled data. Unfortunately, procuring satisfactory training data remains a challenge, thanks to the difficulty and high expense of such processes, especially considering that medical image annotation can only be provided by trained professionals with adequate experience.

In particular, Computed Tomography (CT) and Magnetic Resonance Imaging (MRI) obtain three-dimensional volume data, which requires experts to label each two-dimensional slice image, making manual labeling even more challenging. To tackle these issues, additional algorithms are introduced, including label generation \cite{X-8}, data enhancement \cite{X-9}, and semi-supervised learning schemes that leverage unlabelled data to augment the training data \cite{huang2024flatmatch}. Semi-supervised learning proves to be the more practical approach for various medical image segmentation tasks, where segmentation models are trained with limited labelled data while at the same time benefiting from the abundance of readily available unlabelled data. In general, semi-supervised algorithms utilize two strategies, namely, the pseudo-label generation method and the consistent regularization hypothesis.

Lee et al. \cite{X-10} propose a semi-supervised learning method that uses both labelled and unlabelled data for training.
For unlabelled data, the category with the greatest prediction probability is selected as its pseudo-label for supervised training.
Li et al. \cite{x-11} propose a self-integrating collaborative training framework to automatically extract COVID-19 lesions from CT scans by using limited labelled data and large-scale unlabelled data.
The conservative-aggressive module proposed by Shi et al. \cite{X-14} improves pseudo-label quality by predicting inconsistencies between different misclassification costs and indicating masks for certain and uncertain regions.
Han et al. \cite{X-16} combine the output features of the labelled images of the pre-trained network with corresponding pixel-level annotations to generate class representations according to the mean value operation.
Then pseudo-labels are generated for the unlabelled images by calculating the distance between the unlabelled feature vectors and each class representation,
followed by a series of morphological operations.
Wang et al. \cite{X-17} propose a neighbor matching method,
defining a mapping function that predicts its pseudo-label based on itself and its local manifold.

The consistency assumption is a vital concept in semi-supervised learning.
It posits that if two data points are in close proximity in the input space, their corresponding labels should also be closely related.
In essence, this assumption suggests that similar inputs should produce similar outputs. To this end, Berthelot et al. \cite{X-18} introduce MixMatch, a framework for semi-supervised algorithms.
The algorithm generates various transformations by enhancing unlabelled data and then mandates the model to provide consistent label predictions for these transformations.
Sohn et al. \cite{X-19} propose FixMatch, which simplifies MixMatch.
For every unlabelled sample,
FixMatch conducts powerful and gentle image enhancement, produces unlabelled instances of both powerful and gentle enhanced versions, filters out the generated weak enhanced results using the model's high-confidence threshold, and eventually prompts the model to provide the same forecast for both versions.
The Mutual consistency learning approach proposed by Wu et al. \cite{X-20} compels the model to produce consistent and low-entropy predictions in arduous regions.
It efficiently employs unlabelled data, leverages consistency optimally, and minimizes entropy constraints for model training, ultimately enhancing the efficacy of semi-supervised image segmentation.

\subsection{Tooth Segmentation}
Tooth segmentation involves separating various tooth components, including the crown, root, and pulp, from an oral image. This critical task is necessary for the diagnosis, treatment, and evaluation of teeth in oral medicine. However, due to shape discrepancies, unusual appearance, irregular gum tissue, and insufficient labelling data, tooth segmentation presents a challenging problem. In recent times, tooth segmentation has progressed significantly through the implementation of deep learning methods, predominantly using PXI-produced 2D images or CBCT-produced 3D volumes.

For 2D images, Zhao et al. \cite{X-25} propose a new two-stage attention segmentation network.
This network is not only the first two-stage model for tooth localization and segmentation in panoramic dental X-ray images, but it can also automatically encode rich information and extract more discriminating representations through hierarchical architecture, effectively alleviating the problem of uneven intensity distribution.
The real tooth area can be captured independently of user intervention, while the tooth structure and root topology can be obtained.
Besides, Chen et al. \cite{X-24} propose a novel multi-scale position awareness network that combines multi-scale structural similarity loss, a position awareness module, and an aggregation module to clearly and accurately segment and locate teeth from panoramic X-ray images, thus effectively solving the problem of blurred tooth boundaries.
Cui et al. \cite{X-26} propose a deep learning network that integrates adversarial networks, wide residual blocks, and encoder-decoder structures to learn and extract grayscale and boundary features of teeth under the guidance of a full convolutional network discriminator.
Meanwhile, its loss mechanism effectively avoids network overfitting and enhances feature extraction.

In addition, for 3D images, Cui et al. \cite{X-27} propose a two-level network that first extracts edge maps from CBCT images to enhance image boundary contrast and combines them with original images. Then, a 3D Region Proposal Network (RPN) is constructed with a learning similarity matrix and spatial relationship coding of teeth. This is the first method to segment teeth from CBCT images using a deep learning-based network.
Following that, Rao et al. \cite{X-28} propose a full-convolutional network that combines residual blocks and Dense Conditional Random Field (DCRF) to reduce noise and achieve high-precision segmentation of tooth CBCT images.
Li et al. \cite{X-29} design an attention mechanism to explicitly model semantic graphs of tooth anatomical topological relationships in each quadrant to overcome the segmentation inaccuracies caused by ignoring tooth anatomical topology.
Cui et al. \cite{X-30} propose a two-stage learning framework in which the center of mass and bone marrow are extracted in the first step to determine the approximate position of each tooth, and then a multi-task learning mechanism is used based on the output of the previous step to obtain tooth segmentation results with the help of boundary and root tip information.

It can be seen that most tooth segmentation methods are committed to improving segmentation accuracy and robustness, and the use of neural network technology has achieved impressive results.
Nevertheless, these approaches still have limitations, particularly in terms of the dependence on data annotation, since tooth segmentation necessitates careful manual annotation.
This process is not only time-consuming and labor-intensive but also prone to errors and subjectivity.
Hence, effective learning mechanisms requiring a small quantity of annotated data is highly demanded for tooth segmentation.
As apparent from the preceding discussion, semi-supervised learning is a well established approach that can leverage a significant proportion of unlabelled data in conjunction with a limited amount of labelled data to facilitate network training.
Therefore, we organise this competition, aiming to stimulate new ideas and development of methodologies and solutions that incorporate the semi-supervised learning in tooth segmentation.
The goal of the participating runs is to achieve tooth segmentation results that is equivalent or comparative to those of the fully supervised methods, but by learning from a significantly smaller number of labelled data together with a large quantity of unlabelled data.

\begin{figure*}[!ht]
\centering
\includegraphics[width=\linewidth]{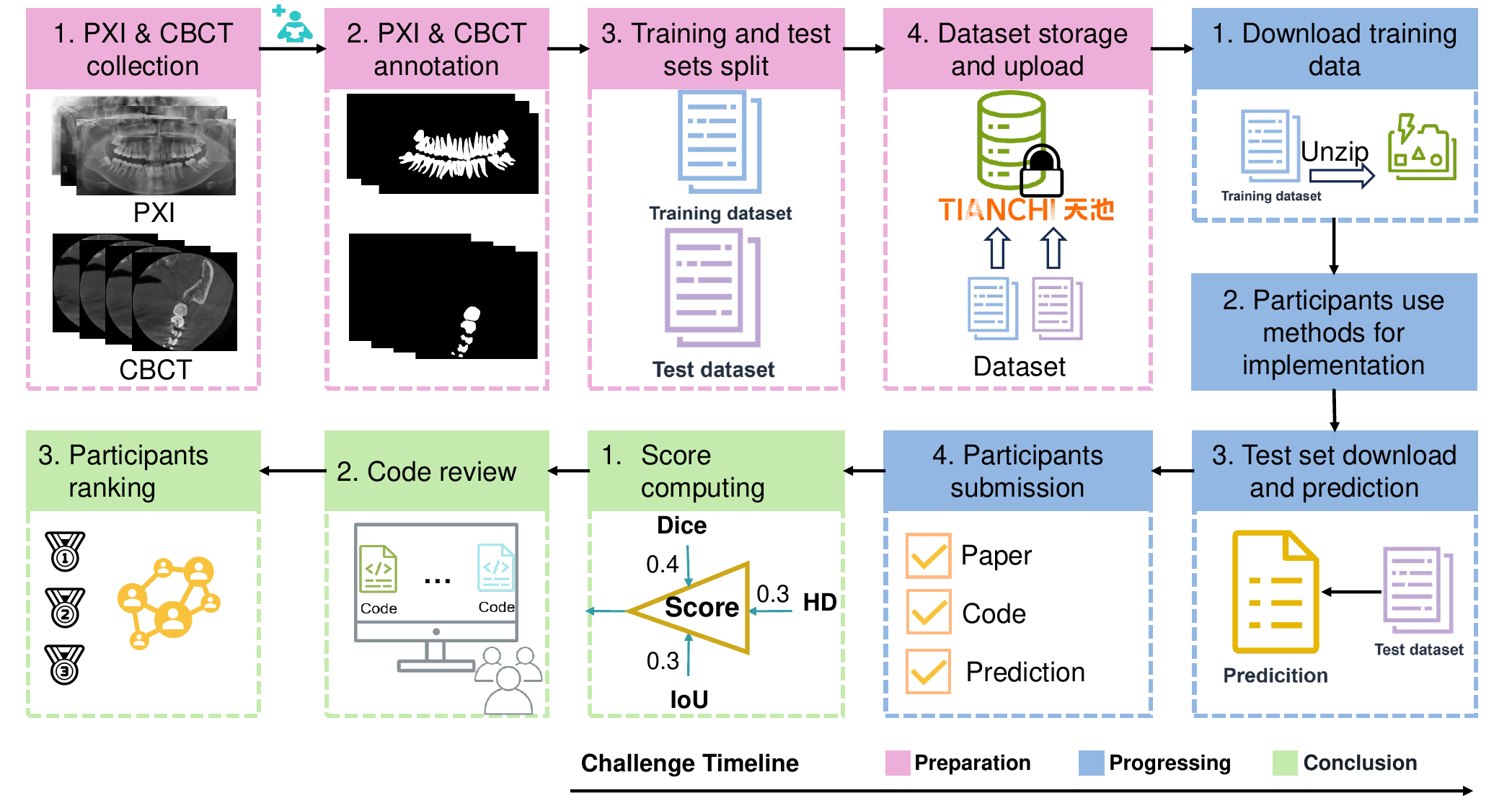}
\caption{The workflow of the Semi-supervised Teeth Segmentation Challenge (STS) challenge consists of three stages: Preparation, Progression, and Conclusion.}
\label{fig:challenges}
\end{figure*}

\section{Challenge Description}
\subsection{Organization}
The objective of the STS challenge is to evaluate the applicability of deep learning image segmentation algorithms on the dental image segmentation tasks, with the special focus on exploring new and robust tooth segmentation methods based on semi-supervision. The STS challenge was conducted as part of the MICCAI 2023 ToothFairy Workshop \footnote{\href{https://toothfairychallenges.github.io/}{https://toothfairychallenges.github.io/}}.
The challenge proposal went through multiple rounds of public review between December 2022 and May 2023. On May 21, 2023, the challenge started with a broad call for participants from all walks of life.
Interested participants can sign up in teams and participate in the challenge through the Alibaba Cloud Tianchi platform.
Members of the organising institutes could participate but are not eligible for awards.
Additionally, the challenge consists of two rounds: the preliminary and the final.
The preliminary round officially began on June 6, and a total of 2,000 dental PXI and 212 CBCT volumes were provided as training data.
For evaluation, a total of 500 dental PXI and 10 CBCT volumes were provided as the test set.
Moreover, the evaluation code was made available to the public before the system was open for submissions \footnote{\href{https://github.com/yefan222/MICCAI-2023-STS}{https://github.com/yefan222/MICCAI-2023-STS}}.
The preliminary round ended on August 27, with the top 100 teams selected for the final round.
The final round started on September 4 and ended on September 15. A total of 3,000 dental PXI and 312 CBCT volumes were provided as training data.
For evaluation, a total of 1,000 dental PXI and 50 CBCT volumes were provided as the test set.
The top seven teams were selected, their codes were collected for review, and the top three teams were selected as the winning teams.
The winners of the first place, the second and third places, were offered \$1,000, \$500, and \$300 monetary prizes, respectively. 

In the final competition, the players were required to containerize their methods through Docker and submit their Docker containers for evaluation on the test set. Only consistent results were considered as valid submissions and kept for ranking. Incomplete submissions that did not include all results were rejected by the evaluation system. 
An example submission was provided to assist participants in verifying the validity of their submissions to the Tianchi platform. Once a submission was successful, the platform automatically updated the players' scores in real time. Each team was allowed to submit three times a day in the preliminary round and two times in the final round to effectively avoid cheating. The workflow of the challenge is shown in Fig. \ref{fig:challenges}.
In this paper, we summarize the main findings and analysis based on the results of the outstanding participants in this challenge; all participants are listed as authors. Following a six-month embargo period, the participating teams can independently publish their own results.

The STS challenge aims to segment the main structure of teeth using two types of dental images: PXI and CBCT.
To accomplish this, the challenge is divided into two tasks: semi-supervised tooth segmentation on panoramic X-ray images (Task 1) and semi-supervised tooth segmentation on cone beam computed tomography volumes (Task 2)
For Task 1, the primary goal is to choose proper training methods to develop a semi-supervised deep learning network model suitable for tooth segmentation, taking as input the panoramic image and the mask corresponding to a part of the original image.
For Task 2, the primary goal is to employ a limited quantity of labelled 3D volume data alongside a substantial amount of unlabelled 3D volume data as inputs for training a 3D network model.
This approach would allow for the successful execution of the task of tooth segmentation within 3D CBCT images utilizing pertinent semi-supervised training techniques and methodologies, despite the minimal amount of labelled data. The target cohort of this challenge include patients who need treatment for common dental diseases (caries, periapical periodontitis, and pulpitis) and orthodontic and endodontic restorative treatment. The treatment of orthodontic and endodontic prosthetic rely on using 3D Cone Beam Computed Tomography scans, while the processes of pinpointing dental disease foci, tooth overall structure captions, and accurate segmentation are often carried out on Panoramic X-rays.

\subsection{Dataset}
Both the two types of data in this challenge were acquired from the intraoral cavity.
For a given patient, one 2D/3D scan was acquired, covering the full or part of
the upper and lower jaws areas with teeth. The data was collected for patients requiring either preoperative examination or prosthetic treatment.
The proposed computational methods were expected to segment all visible teeth in a given 2D or 3D scan.  All dental scans were categorized into four classes, including missing teeth with appliance, missing teeth without appliance, teeth with appliance and teeth without appliance.
The dataset was built with an even distribution among all cases. The scans were captured by orthodontists with more than five years of professional experience.
Then, they were annotated by fifteen dentists. Twelve junior dentists with at least two years of experience first delineated tooth regions slice-by-slice in the axial view, and modified the annotations in the coronal view and sagittal view. Then, the senior dental experts assessed the annotation quality, and marked a quality level (in excellent, good, fail and
poor) on each tooth annotation. “Excellent” annotations were stored in the dataset directly. “Good”
annotations were fine-tuning according to the experts’ feedback. “Fair” and “Poor” annotations and their feedback
were put back into the unlabelled data pool and were marked again, until the annotation quality achieves the required level. 
Ethical approval of the data used in this challenge was obtained from the Medical Ethics Committee of Sichuan Provincial People's Hospital and the University of Electronic Science and Technology Hospital Research Ethics Committee (No. 2022YR014). The data usage agreement license is under CC BY-NC-ND.
\begin{table*}[!ht] 
\centering
\caption{The dataset information for the PXI semi-supervised segmentation task in the STS challenge, including the training and testing splits, and the proportion of labelled and unlabelled data.}
\label{tab:dataset2d}
\setlength{\tabcolsep}{6mm}      
\renewcommand\arraystretch{1.3}  
\begin{tabular}{lcccccc} 
\hline
\toprule
\multirow{2}{*}{Stage} & \multirow{2}{*}{Class} & \multicolumn{2}{c}{Train} & \multicolumn{2}{c}{Test} & \multirow{2}{*}{Total}  \\ 
\cmidrule(r){3-4} \cmidrule(r){5-6}
        &     & unlabelled & labelled      & unlabelled & labelled     &             \\ \hline
\multirow{3}{*}{Preliminary round} & Children        & -----         & 150            & -----         & 300           & 450       \\ 
                                   & Adult           & -----         & 1850           & -----         & 200           & 2050                    \\ 
                                   & Total           & -----         & 2000           & -----         & 500           & 2500                    \\ 
\hline
\multirow{3}{*}{Final round}       & Children        & 100           & 50             & -----         & 350           & 500              \\ 
                                   & Adult           & 2000          & 850            & -----         & 650           & 3500             \\ 
                                   & Total           & 2100          & 900            & -----         & 1000          & 4000                \\ 
\hline
\toprule
\end{tabular}
\end{table*}

\begin{table*}[ht]
\centering
\caption{The dataset information for the CBCT semi-supervised segmentation task in the STS challenge, including the training and testing splits, and the proportion of labelled and unlabelled data.}
\label{tab:dataset3d}
\setlength{\tabcolsep}{8mm}      
\renewcommand\arraystretch{1.3}  
\begin{tabular}{lccccc} 
\hline
\toprule
\multirow{2}{*}{Stage}  & \multicolumn{2}{c}{Train} & \multicolumn{2}{c}{Test} & \multirow{2}{*}{Total}  \\ 
\cmidrule(r){2-3} \cmidrule(r){4-5}
                    & unlabelled & labelled        & unlabelled  & labelled     &             \\ \hline
Preliminary round   & 200       & 12             & -----      & 10          & 222         \\ 
Final round         & 300       & 12             & -----      & 50          & 362         \\
\hline
\toprule
\end{tabular}
\end{table*}

\subsubsection{2D-PXI Dataset}
The 2D-PXI dataset was mainly obtained from Hangzhou Dental Hospital and Hangzhou QianTang Dental Hospital, followed by the dental image dataset that has been published by \cite{Y-12}. According to the age of the patient and the tooth morphology presented, all the panoramic tooth images are divided into: adult panoramic tooth images (AD-PXI) and child panoramic tooth images (CD-PXI).  During the pre-processing phase, we converted the image format from DICOM to PNG. Dental experts from the hospital annotated the data using EISeg \cite{liu2021paddleseg}, LableMe \cite{russell2008labelme} and other tools. We then obtained a three-channel 24-bit grayscale images with a pixel value of 640 $\times$ 320 and the corresponding binary masks. Some statistical information of the 2D-PXI dataset can be seen in Table \ref{tab:dataset2d}. When conducting dataset partitioning, we uniformly allocated the training, validation, and test sets according to the real ratio of adult to pediatric patients in the clinic to avoid any bias in data distribution in both preliminary and final round.


\subsubsection{3D-CBCT Dataset}
The 3D-CBCT dataset was mainly provided by Hangzhou Qiantang Dental Hospital, followed by a publicly available dataset CTooth  \cite{X-43,X-44}. For the annotation work of each volume, dentists in the hospital first marked the tooth area layer by layer in the axial direction with the help of ITK-SNAP \cite{yushkevich2006user} software, and then corrected and annotated the tooth area with the help of coronal and sagittal perspectives. Additional dentists were invited to evaluate the above annotation work and correct the poor evaluation results. These dental CBCT images were taken using an OP300 manufactured with the Instrumentarium Orthopantomograph$^\circledR$. All dental CBCT slices were scanned before dental surgery, with an axial resolution of 640 $\times$ 640 pixels, a slice thickness of 0.25mm, and a total thickness of 99.75mm. Some statistical information of the 3D-CBCT dataset can be seen in Table \ref{tab:dataset3d}. 


\subsection{Performance Evaluation}
In the competition, all participants submit the predicted segmentation masks on the provided original test images, packaged and uploaded in .png (2D PXI segmentation task) or .nii.gz (3D CBCT segmentation task) formats.
In the competition, three indicators are applied for quantitative measurement of the segmentation performance: Dice coefficient, Mean Intersection over Union (mIoU) and Hausdorff Distance (HD). The three indicators are described in detail below.

The Dice coefficient is a set similarity measure function that evaluates the similarity of two sets and can be formalized as follows:
\begin{align}
    \text{Dice} = \frac{2 *|A \cap B|}{|A|+|B|}.
\end{align}
where $A$ represents the mask predicted by the proposed model, and $B$ represents the actual mask of Ground Truth (GT).

The Mean Intersection over Union (mIoU) is used to measure the ratio of the intersection over the concatenation of two sets, namely, the GT and predicted masks. It can be defined as follows:
\begin{align}
    \text{mIoU} = \frac{1}{k} \sum_{i=1}^{k} \frac{T P}{F N+F P+T P}.
\end{align}
where $k$ is the number of classes (including empty classes, i.e. background), $T P$, $F N$, and $F P$ indicate true positive, false negative, and false positive, respectively.

Hausdorff distance (HD) is the minimum distance between two shapes or curves obtained by the Hausdorff transformation. In the 2D PXI segmentation task, the 2D HD formula is defined as follows:
\begin{align}
    \text{HD}_{2d} = \min\big(|x_{1} - x_{2}| + |y_{1} - y_{2}|\big).
\end{align}
where $(x1, y1)$ and $(x2, y2)$ represent the coordinates of the two pixels; $| x_{1}-x_{2} |$ and $| y_{1}-y_{2} |$ represent the distances on the corresponding axes. This formula represents the sum of the absolute distances between two pixels in a two-dimensional medical image on the horizontal and vertical axes, which is the HD at the pixel level.

In the 3D CBCT segmentation task, the 3D HD formula is as follows:
\begin{align}
    \text{HD}_{3d} = \min\big(|x_{1} - x_{2}| + |y_{1} - y_{2}| + |z_{1} - z_{2}|\big).
\end{align}
where $(x_{1}, y_{1}, z_{1})$ and $(x_{2}, y_{2}, z_{2})$ represent the coordinates of the two voxels; $| x_{1}-x_{2} |, | y_{1}-y_{2} |$, and $| z_{1}-z_{2} |$ represent the distances on the corresponding axes. This formula is a voxel-level measure of distance. It is worth noting that after calculating the above two HDs, both of them are normalized so that the final score can be calculated.

Finally, each competitor's overall score  is calculated as the weighted average of the three evaluation criteria presented above, using the following:
\begin{align}
    \text{Score} = 0.4 \ast \text{Dice} + 0.3 \ast \text{mIoU} + 0.3 \ast \big( 1-\text{HD} \big). 
\end{align}

We consider these evaluation metrics of similar importance. Therefore, this simple weighting scheme is determined empirically and is expected to lead to a fair ranking overall. In addition, we provide intermediate rankings based on each metric, in order to highlight the outstanding performances in the specific aspects. 
Ranking variability is characterized using the bootstrap method.

\section{Challenge Submissions}

\begin{table*}[!ht]
\centering
\caption{Quantitative evaluation results of the top 10 qualified teams in terms of (mean ± standard deviation) Dice Similarity Coefficient (Dice), mean Inter over Union (mIoU) and Hausdorff Distance (HD). The top three scores in each metric are highlighted in \textbf{bold}.}
\label{tab:2d3dperform}
\setlength{\tabcolsep}{2.5mm}      
\renewcommand\arraystretch{1.3}    
\begin{tabular}{llll|llll} 
\hline
\toprule
PXI Teams & Dice                & mIoU                & HD                     & CBCT  Teams & Dice                & mIoU                & HD                      \\ 
\hline
T1    & \textbf{93.92±4.99} & \textbf{98.30±0.94} & 0.0272±0.0439          & T1    & \textbf{84.42±5.41} & \textbf{86.61±3.84} & \textbf{0.1595±0.1104}  \\ 
T2    & \textbf{93.71±3.93} & \textbf{98.23±0.91} & 0.0239±0.0351          & T2    & \textbf{83.43±4.93} & \textbf{85.83±3.49} & 0.1615±0.1121           \\ 
T3    & 93.57±5.02          & \textbf{98.19±1.19} & \textbf{0.0224±0.0346} & T3    & 80.58±5.02          & 83.76±3.36          & \textbf{0.1599±0.1115}  \\ 
T4    & \textbf{93.65±2.52} & 98.16±0.79          & \textbf{0.0232±0.0168} & T4    & 80.70±4.95          & 83.85±3.37          & 0.1689±0.1052           \\ 
T5    & 93.41±5.07          & 98.14±1.04          & 0.0244±0.0455          & T5    & 79.31±5.67          & 82.90±3.75          & 0.1708±0.0959           \\ 
T6    & 93.34±5.00          & 98.12±0.87          & 0.0239±0.0451          & T6    & \textbf{81.23±6.49} & \textbf{84.31±4.17} & 0.2344±0.0456           \\ 
T7    & 93.28±3.90          & 98.09±0.80          & 0.0243±0.0342          & T7    & 77.49±4.82          & 81.61±3.09          & \textbf{0.1580±0.1080}  \\ 
T8    & 93.16±4.64          & 98.09±0.83          & 0.0245±0.0352          & T8    & 78.65±5.59          & 82.43±3.62          & 0.1844±0.0917           \\ 
T9    & 93.00±2.85          & 97.97±1.01          & \textbf{0.0234±0.0159} & T9    & 76.64±6.47          & 81.11±3.96          & 0.1870±0.0933           \\ 
T10   & 92.99±4.05          & 98.00±0.95          & 0.0243±0.0367          & T10   & 78.82±6.36          & 82.60±4.05          & 0.2413±0.0477           \\
\hline
\toprule
\end{tabular}
\end{table*}
In the preliminary round, Task 1 and Task 2 received the submissions from 373 and 61 teams, respectively. Among these, 44 and 20 teams qualified for the final round for the respective tasks.

\subsection{Submitted methods in Task 1: STS on 2D-PXI dataset}
The overall benchmark of the top teams can be seen in Table \ref{tab:top10_PXI_summary}.
In the following, we provide a concise description of the methods that led to the top scores, highlighting the technical key points. 

\textbf{Rank 1:} Zhuang proposed an adaptive fusion network for tooth region segmentation that trains two groups of teacher-student models simultaneously based on the Mean-Teacher model \cite{X-21}.
First, the method used 900 PXI with masks, increases the amount of data by reducing the slide step size (adjusted from 32 to 16), and trained two teacher models based on different network architectures, the backbone networks being SegResNet\_64 \cite{Y-6} and DynUNet \cite{Y-6}.
The two teacher models then fused to make predictions on 2,100 unlabelled images, generating 3,000 soft labels.
These soft labels took full advantage of both teacher models.
Next, using these 3000 newly generated soft labelled images as training data,
two student models with a similar structure were trained separately.
Finally, all the teacher and student models were fused into one powerful model that predicts the results. For the loss function, the teacher model was trained with BCEWithLogitsLoss.
Besides, MSELoss was used when the student model was trained with soft labels.
The prediction by the student model was made as close as possible to the soft label by the teacher model, with the aim to learn the ability mastered by the teacher model.
AdamW was used in the optimizer section, with step size set to 3e-3 and weight decay set to 1e-5. For the trained models, data enhancement was added. Each image was rotated clockwise and features were obtained on the rotated versions, and then these features were reversely rotated and merged, obtaining a final feature map by taking the mean in the channel dimension.
In summary, Zhuang's submission achieved the highest segmentation performance with an overall score of 0.9624, Dice of 0.9392, mIoU of 0.9830, and HD of 0.0272.

\textbf{Rank 3:} Liu et al. performed data cleaning on the training set in the pre-processing stage to screen out data with more accurate labelling and used the 5-fold division method to re-divide the new training data into training and validation sets (training :validation = 4:1).
In addition, they used data enhancement strategies such as flipping, rotating, mixing, and coarse dropout. Two types of segmentation networks were used in the training process: a U-Net \cite{X-1} like model with EfficientNetv2 \cite{Y-7} as the backbone feature extraction network and a model with SAM \cite{Y-8} as the baseline feature extraction network.
The UNet-like EfficientNetv2 model was fine-tuned based on the pre-training weights from ImageNet, and the SAM model was fine-tuned based on the pre-training weights on the SA1B dataset.
For the specific training process, the model was first trained with labelled data, and then the model was used to predict unlabelled data to get the corresponding pseudo-masks.
900 pairs of images and masks were randomly selected and added to the training set for the next round of training.
In terms of data, two different image resolutions were used: one is 1024 $\times$ 1024 and the other is 640 $\times$ 1280. In addition, three loss functions were used in the optimization process: Cross-entropy loss, Dice loss, and Lovasz loss. The optimizer used AdamW, and the cosine learning rate strategy was chosen for the learning rate strategy.

\textbf{Rank 5:} Wang et al. proposed an improved model based on UperNet \cite{Y-9}. The backbone network was replaced with ConvNeXt \cite{Y-10}.
They also introduced a class imbalance loss, where the foreground has a larger weight and suffers a greater penalty when the model forecasts incorrectly.
Additionally, online hard sample mining was implemented, meaning that only pixel value points with confidence scores of 0.9 or higher will be used for training, with a minimum of 100,000 pixel value points retained during the process. 
To enhance the diversity of the datasets, mosaic data augmentation and multi-scale scaling training techniques were employed.
Furthermore, The test time augmentation method was used for inference and a post-processing method was proposed.
The post-processing method involved generating an inference result probability matrix for each image during inference, with dimensions of (2, 320, 640). Each pixel point within the matrix was assigned two probabilities: one for the tooth region and one for the non-tooth region, respectively.
This matrix was obtained by traversing all the pixel points in the image.
If the maximum value between the two predicted probabilities for a pixel point was less than 0.6, it indicated that the model was unable to accurately distinguish the category to which the pixel belongs. In such cases, the pixel value of the point was changed to represent the background, while the rest of the points were assigned to the tooth region.

\textbf{Rank 7:} Lin et al. proposed an optional retrained, semi-supervised tooth segmentation model.
The training of the model is divided into three phases, which are the easy label prediction phase, the hard label re-generation phase, and the final model training phase.
Specifically, they first designed a segmentation model with a denoiser; both denoiser and segmentation models are U-Net-based architectures \cite{X-1}, and the denoiser and segmentation model were connected in series.
The denoiser accepted the segmentation figures generated by the segmentation model as an input, outputted the noise in the segmentation label.
Finally, the segmentation label will be subtracted from the noise to get the segmentation result after noise reduction.
This strategy improved the quality of the segmentation results and reduced the noise of the pseudo-labels at the same time.
Their training phase was divided into three stages.
The first stage used only labelled data for fully supervised training and saved checkpoints at three training nodes to generate three sets of pseudo-labels for unlabelled data, respectively.
The mean Dice scores of the three sets of pseudo-labels were then calculated and sorted from high to low, and the top 75\% of pseudo-labels were selected to join the second stage of training.
Strong data augmentation and hard data augmentation (Mixup and CutMix) were performed for unlabelled data, and only weak data augmentation was used for labelled data.
Then pseudo-labels were generated for the remaining 25\% of unlabelled data using the model trained in the second phase.
Finally, the third stage used labelled data with all unlabelled data for training.

\textbf{Rank 8:} Song et al. chose FCBformer \cite{Y-5} as the main model architecture.
FCBformer skillfully combined two key techniques, Fully Convolutional Networks (FCNs) and Transformers, by starting the processing of downsampled images through two parallel branches and then concatenating the output tensors of these two branches and the output tensors of FCNs in the channel dimension.
Finally, the prediction head processed the concatenated tensor to generate a full-size segmentation map of the input image.
In terms of training strategy, they adopted the Exponential Moving Average (EMA) strategy for smoothing the model parameters to stabilize the training process and enhance the generalization ability of the model.
At the same time, the learning rate and decay factor were adjusted to reduce the model error on the validation set while maintaining the training speed.
The loss function was a combination of cross-entropy loss and dice loss, which was designed to balance the accuracy of pixel-level classification and the consistency of segmentation regions as a way to improve the model's performance on tooth segmentation tasks.
In addition to this, they used the Lion optimizer \cite{Y-6} for model optimization.
In the final round of the challenge,
they used the model obtained from the training in the preliminary phase as a pre-training model to accelerate the convergence of the final model and to improve the performance of the final model by leveraging the knowledge that the preliminary model has already learned.
The application of the pre-training model not only shortened the model training time but also improved the accuracy of the model on the rematch task.
This series of innovative strategies enabled their model to achieve excellent results on the tooth segmentation task, demonstrating the strong potential of deep learning in the field of medical image analysis.

\begin{figure*}[!ht]
  \centering
  \begin{subfigure}{0.31\textwidth}
    \centering
    \includegraphics[width=\textwidth]{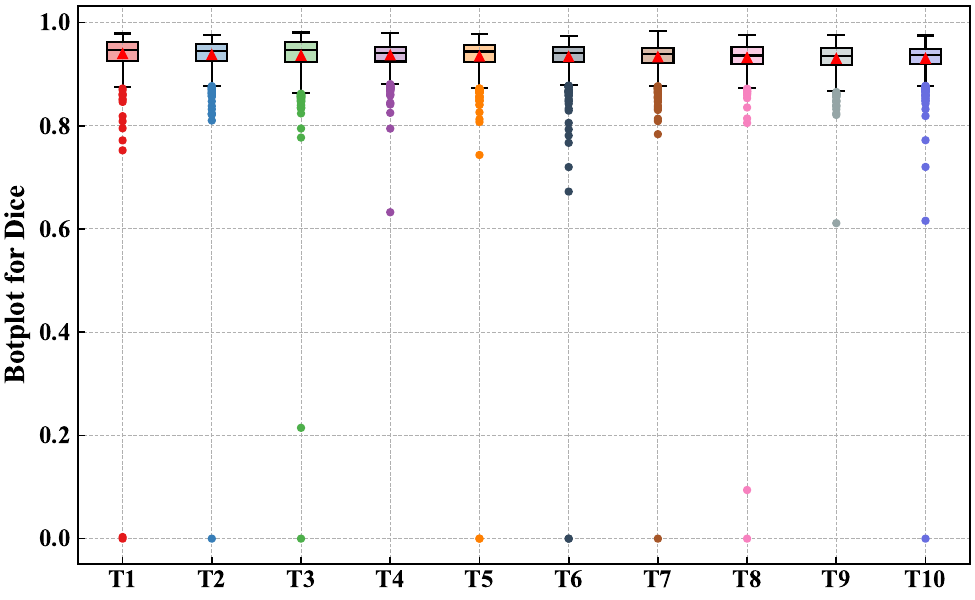}
    \caption{}
  \end{subfigure}
  \hfill
  \begin{subfigure}{0.31\textwidth}
    \centering
    \includegraphics[width=\textwidth]{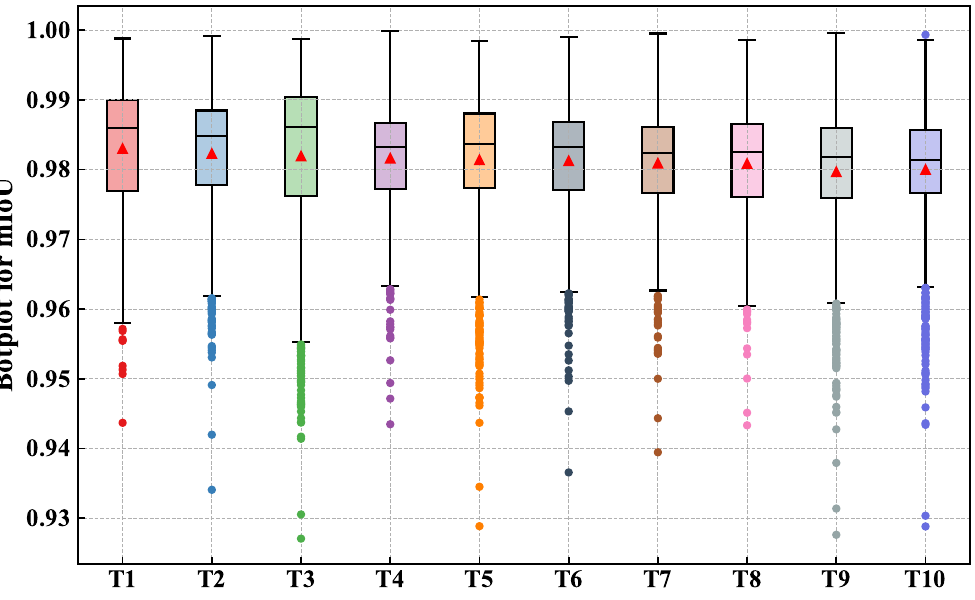}
    \caption{}
  \end{subfigure}
  \hfill
  \begin{subfigure}{0.31\textwidth}
    \centering
    \includegraphics[width=\textwidth]{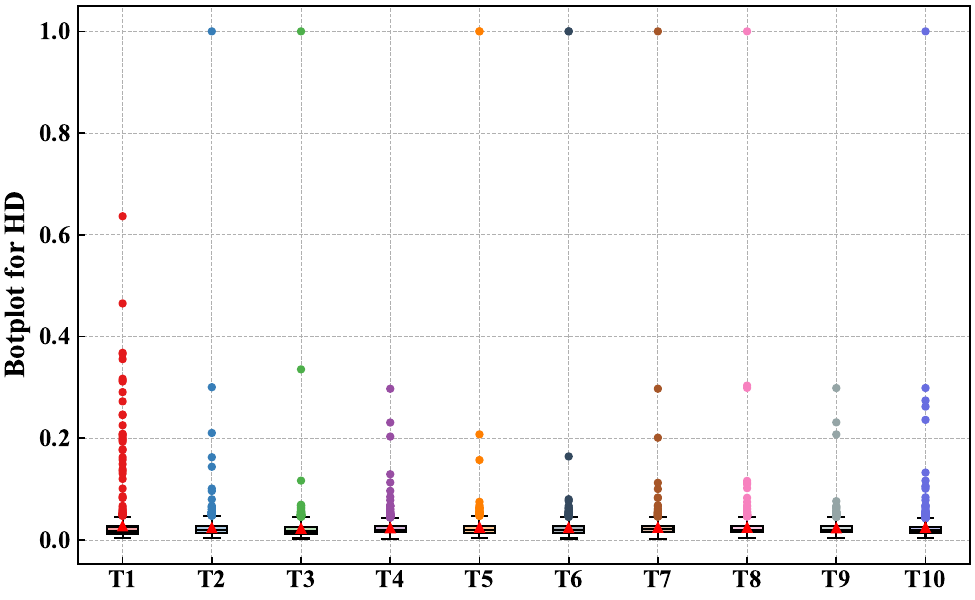}
    \caption{}
  \end{subfigure}
  \hfill
  \begin{subfigure}{0.31\textwidth}
    \centering
    \includegraphics[width=\textwidth]{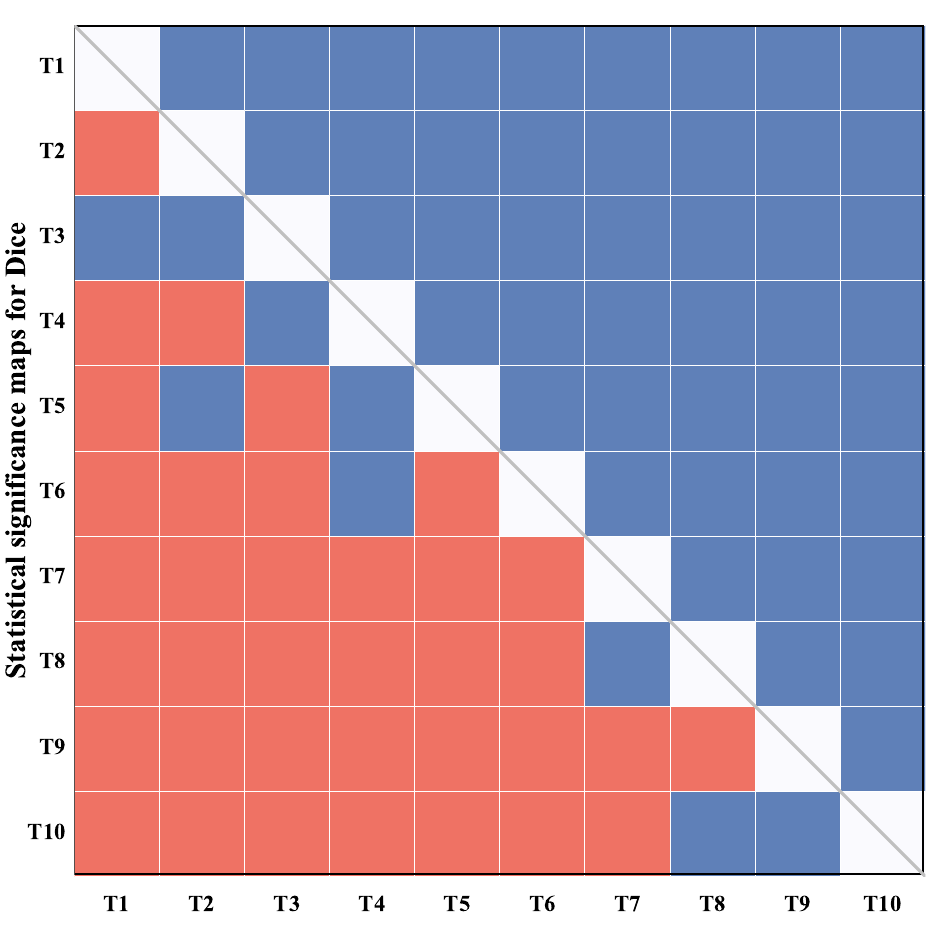}
    \caption{}
  \end{subfigure}
  \hfill
  \begin{subfigure}{0.31\textwidth}
    \centering
    \includegraphics[width=\textwidth]{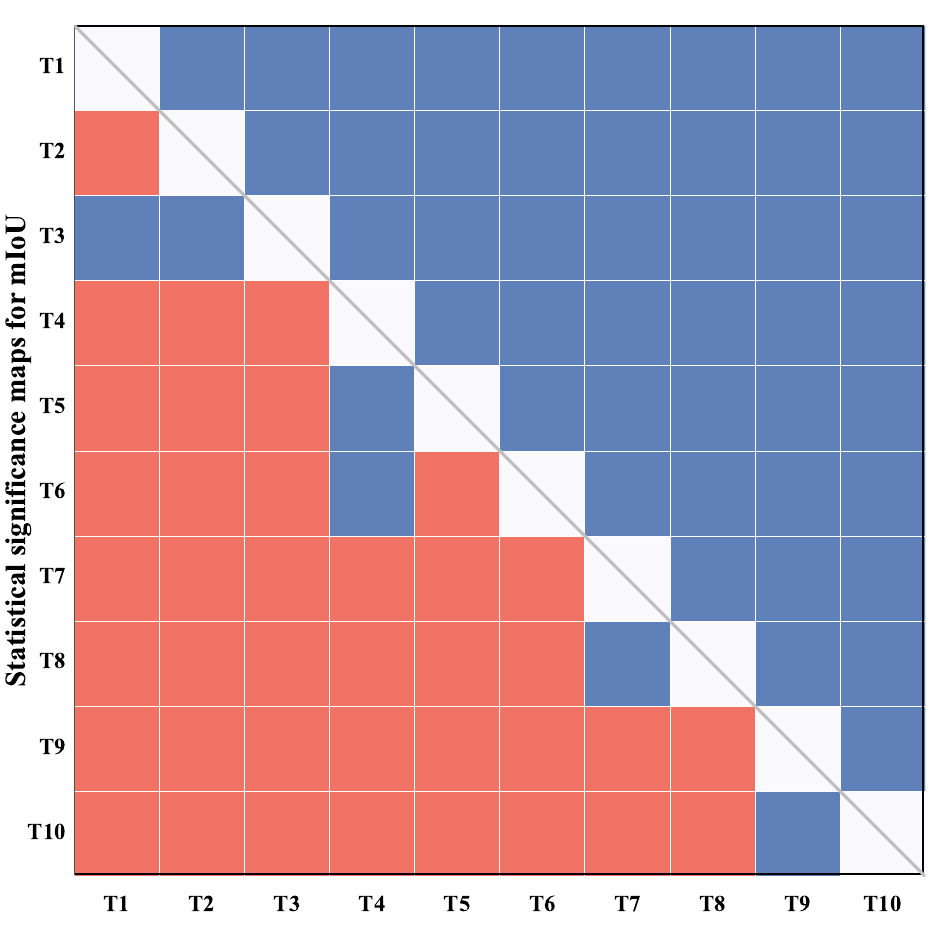}
    \caption{}
  \end{subfigure}
  \hfill
  \begin{subfigure}{0.31\textwidth}
    \centering
    \includegraphics[width=\textwidth]{images/pvalue_iou2d.pdf}
    \caption{}
  \end{subfigure}
  \caption{Boxplot visualization (the first row) and statistical significance maps (the second row) for 2D tooth segmentation track. (a) and (d), (b) and (e), and (c) and (f) are the results for the Dice, mIoU and HD, respectively.}
  \label{fig:result2d}
\end{figure*}

\textbf{Rank 11:} Xue et al. used automatic generation alongside conventional data augmentation to expand the dataset and enhancing model generalization.
Specifically, for each tooth in the image, they randomly used two generation methods, i.e., removing all of the individual teeth and randomly scaling the morphology of individual teeth.
This generation method was well suited to characterize the data in this task domain while solving the problem of significant differences in the distribution of samples with small data volumes.
The network part used DecoupledSegNets \cite{Y-11} as the backbone.
The model architecture was characterized by splitting the foreground into the edge part (high frequency) and the body part (low frequency), which were supervised to learn separately and then merged, aligning with the data characteristics.
They also extracted branches at different depths of the model and utilized dynamic mutual loss for additional supervision.
Besides, they divided the prediction process into coarse and fine segmentation.
Both segmentation processes used the same network structure as described above.
The input image is initially segmented by the coarse model.
Next, the maximum outer join matrix was applied to remove excess background information.
Subsequently, the fine model executes the fine segmentation process.
The data from the training process was processed in the same way as above.

\subsection{Methods of Task2: STS on 3D-CBCT dataset}
The overall benchmark of the top teams can be seen in Table \ref{tab:top10_CBCT_summary}. Similarly, we list the key points of the participating teams who provided their key strategies in the 3D CBCT dental semi-supervised segmentation. The details are presented as below.

\textbf{Rank 1:} 
Li et al. proposed a multi-stage training strategy to address the challenges of CBCT image segmentation. The key to this strategy is to decompose the task into three sub-tasks and train the model step by step to better capture the contextual information of the image.
They extended the properties of nnUNet \cite{X-7} by introducing additional layers into the decoder of nnUNet to increase the depth of the model and increase the number of channels in the neural network.
Based on these modifications, they adapted the nnUNet served as the backbone of the entire algorithm.
In the three-stage model training strategy,
the first stage consisted of constructing a basic dental model using the given labelled data.
It is observed that bone segmentation can be achieved by combining the basic tooth model through a simple thresholding method.
In order to accurately distinguish teeth, maxilla, and mandibles,
a second stage of processing was introduced to specialize in a subnetwork of jaw segmentation.
Finally, a third stage is combined with jaw information to segment teeth.
In the first stage,
a preprocessed dataset was used to train the nnUNet,
and pseudo-labels were generated for a large amount of unlabelled data.
The generated pseudo-labels \cite{X-32} were automatically and manually selected, optimized and filtered, and then incorporated into the training iterations to form a high-quality training dataset.

\textbf{Rank 2:} Wang et al. proposed a semi-supervised algorithm based on the nnUNet network \cite{X-7} and pseudo-labeling strategy \cite{X-33}.
Firstly, the labelled 3D data were preprocessed by morphological opening and closing operations and connectivity domain analysis, and then fed into the nnUNet network for segmentation.
The unlabelled 3D data were also pre-processed as described above and fed into the network for prediction, and the resulting segmentation results were filtered and used as pseudo-labels for subsequent training, while pseudo-labels were screened for uncertainty detection.
After each training epoch, the training dataset was continuously expanded to train the subsequent network. The loss function combined dice loss and cross-entropy loss, and the initial learning rate was set to 0.01, and the number of training rounds was set to 1000.

\begin{figure*}[!ht]
  \centering
  \begin{subfigure}{0.31\textwidth}
    \centering
    \includegraphics[width=\textwidth]{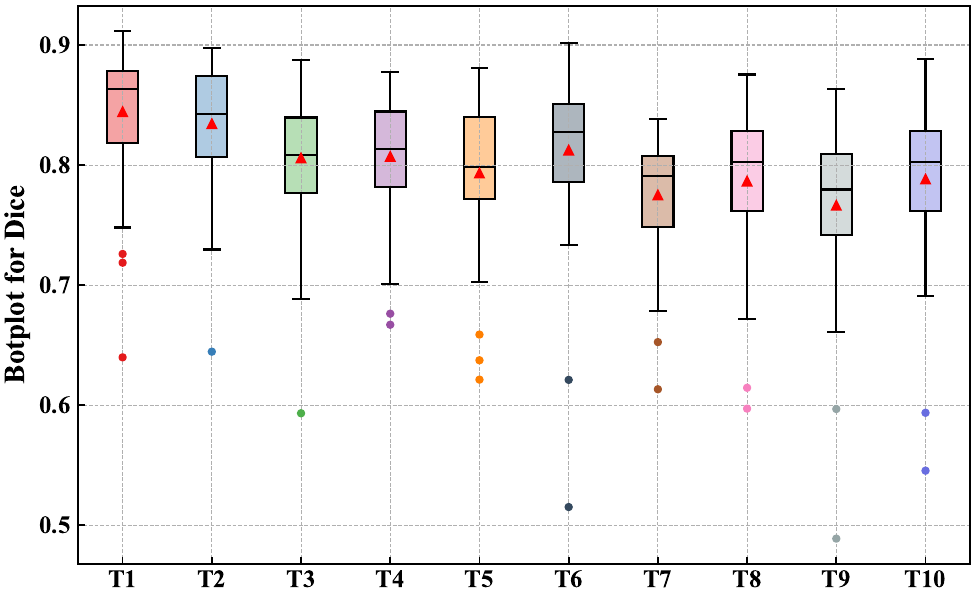}
    \caption{}
  \end{subfigure}
  \hfill
  \begin{subfigure}{0.31\textwidth}
    \centering
    \includegraphics[width=\textwidth]{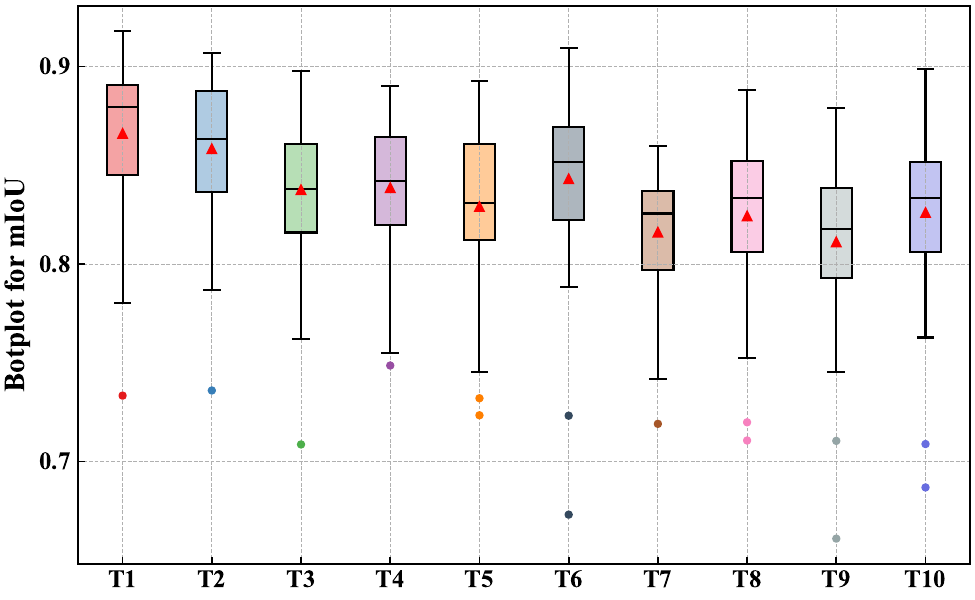}
    \caption{}
  \end{subfigure}
  \hfill
  \begin{subfigure}{0.31\textwidth}
    \centering
    \includegraphics[width=\textwidth]{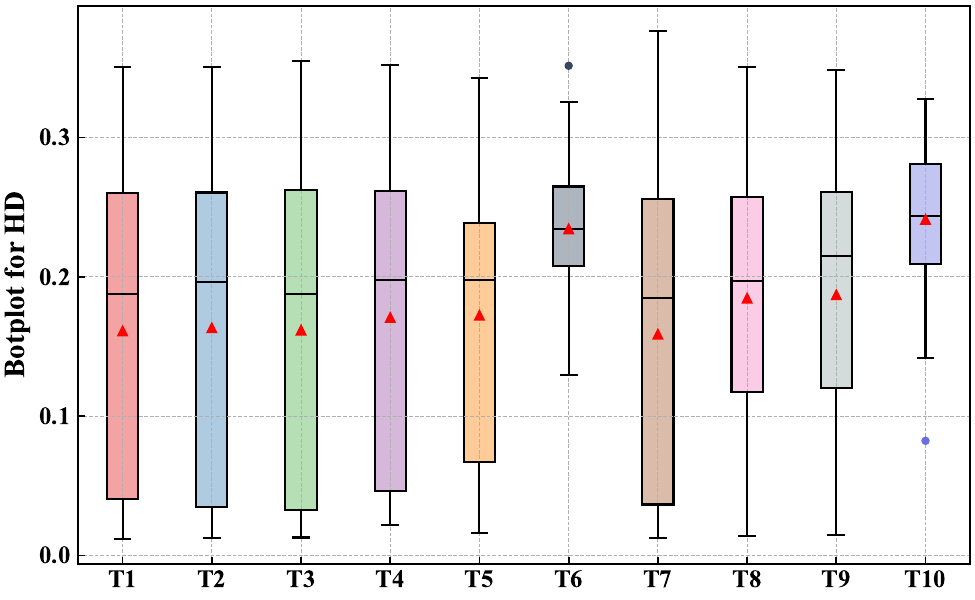}
    \caption{}
  \end{subfigure}
  \hfill
  \begin{subfigure}{0.31\textwidth}
    \centering
    \includegraphics[width=\textwidth]{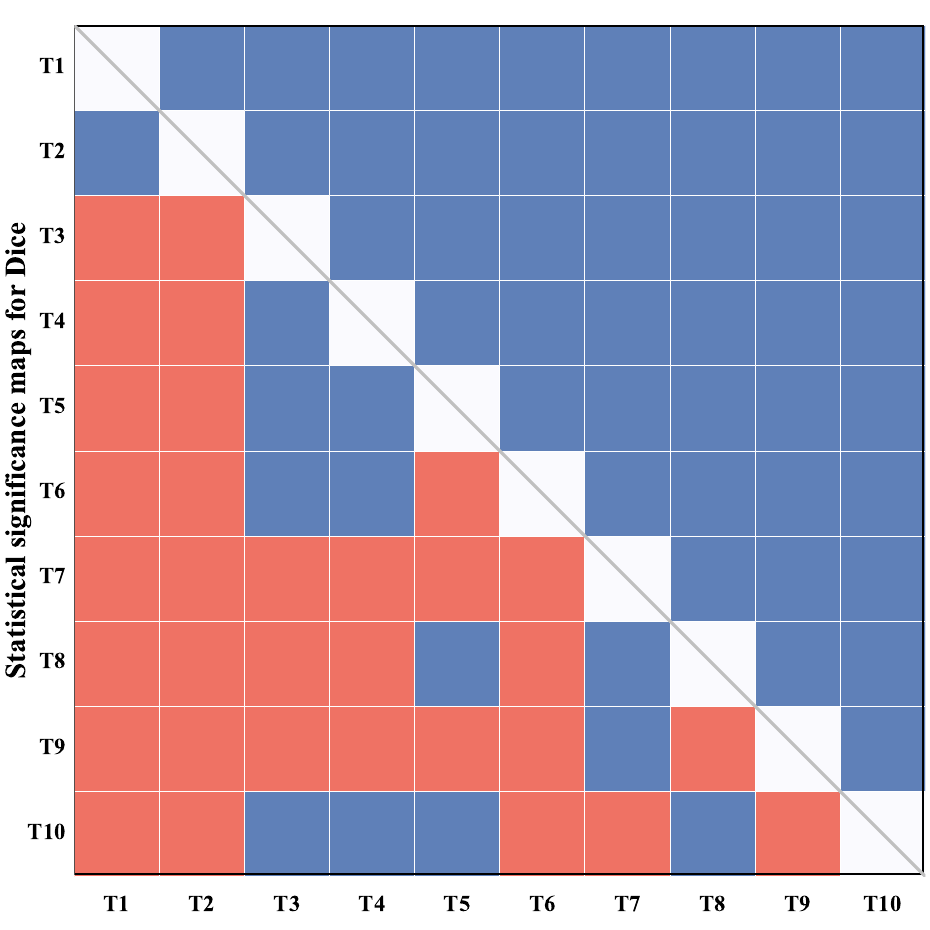}
    \caption{}
  \end{subfigure}
  \hfill
  \begin{subfigure}{0.31\textwidth}
    \centering
    \includegraphics[width=\textwidth]{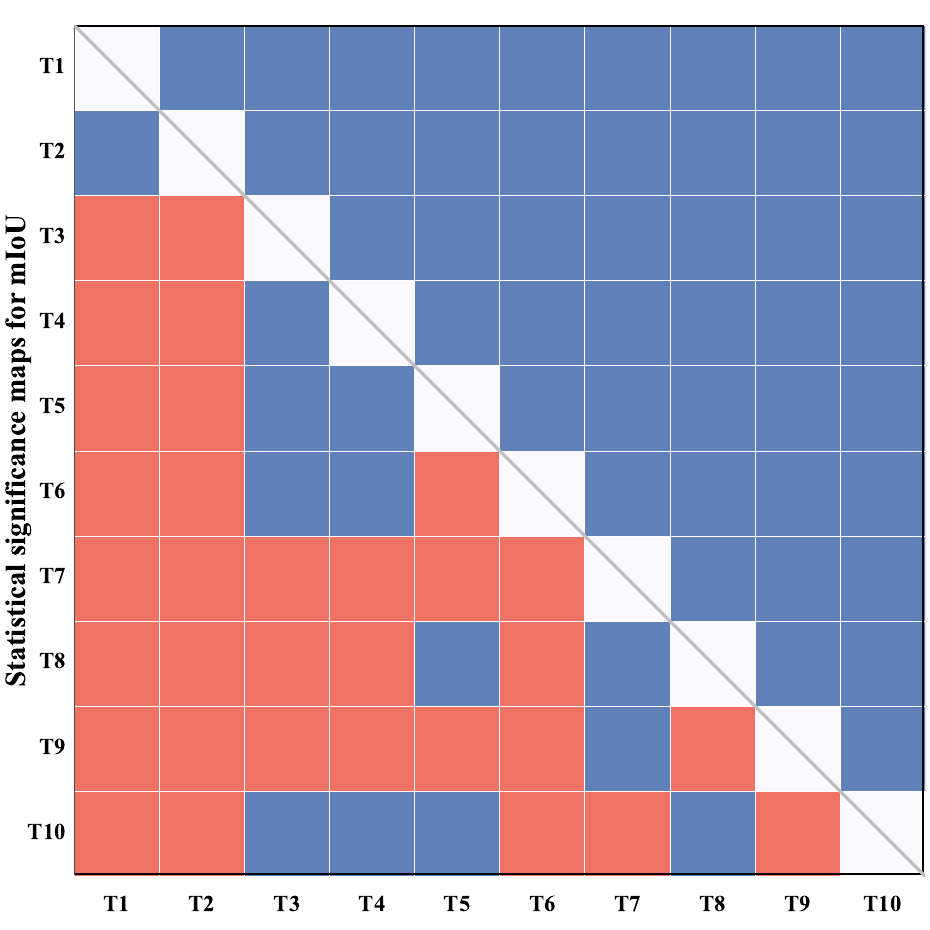}
    \caption{}
  \end{subfigure}
  \hfill
  \begin{subfigure}{0.31\textwidth}
    \centering
    \includegraphics[width=\textwidth]{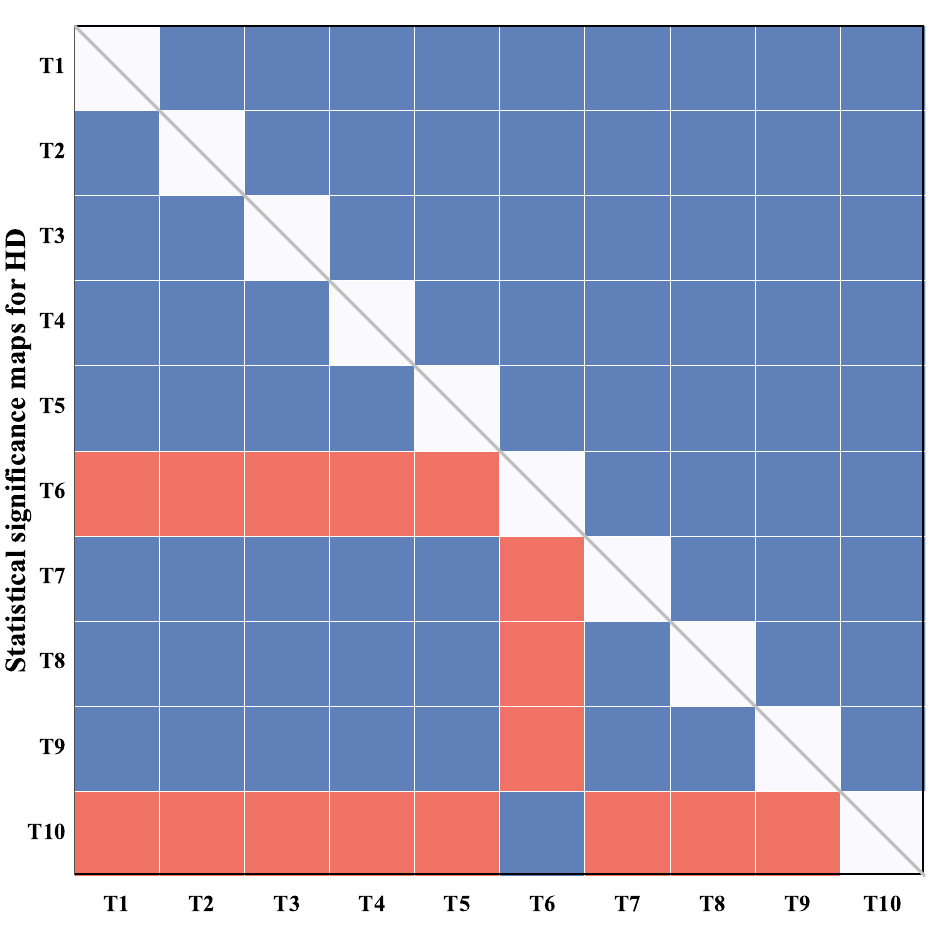}
    \caption{}
  \end{subfigure}
  \caption{Boxplot visualization (the first row) and statistical significance maps (the second row) for 3D tooth segmentation track. (a) and (d), (b) and (e), and (c) and (f) are the results for the Dice, mIoU and HD, respectively.}
  \label{fig:result3d}
\end{figure*}

\textbf{Rank 3:} Wang et al. proposed a framework for a semi-supervised multi-stage training algorithm based on a Fourier Transform Augmentation (FTA) data augmentation module and an improved UniMatch \cite{X-34}.
The FTA data augmentation module was developed by Fourier transforming randomly selected piece of labelled data and a piece of unlabelled data, merging them in the frequency domain space \cite{X-35}, and then performing Fourier inverse transformed to obtain the respective augmented images.
The improved UniMatch framework introduced the adaptive adjustment of the confidence threshold in FreeMatch \cite{X-36} on the basis of the UniMatch algorithm framework.
In the multi-stage training framework, Wang et al. first used a 2D nnU-Net model for supervised learning to generate pixel-level pseudo-labels for a small number of randomly sampled unlabelled samples in the training set.
Then, the data processed in the first stage was subjected to data enhancement by the FTA module, which enables the model to learn information on the target domain, and then the improved UniMatch model was used for feature learning.

\textbf{Rank 4:} Liu et al. proposed a semi-supervised noise-resistant segmentation network and a three-stage training framework.
The noise-resistant semi-supervised framework mainly used two parallel networks so that they can learn from each other.
In addition, Liu et al. divided the dataset into two mutually exclusive subsets, and these different subsets were fed into two different networks to ensure that the networks learn more diverse features.
Both networks used noisy labels for initial supervision, and their respective predictions were fused as further pseudo-labels, and the weight of these pseudo-labels was gradually increased.
They also used the noise transfer matrix \cite{X-38,X-39} to correct the loss of noisily labelled tooth edge regions.
In a three-stage training framework, the first stage used the traditional semi-supervised segmentation method Mean-Teacher \cite{X-21} to obtain preliminary full-field segmentation results.
The second stage merged the unlabelled data from the first stage with the generated pseudo-labels into the labelled dataset and then feeded the above data into the constructed noise-resistant semi-supervised framework to obtain more accurate background region segmentation results.
In the third stage, only the regions of interest were cropped and fed into the noise-resistant semi-supervised network to achieve finer segmentation.

\begin{figure*}[!t]
\centering
\includegraphics[width=\textwidth]{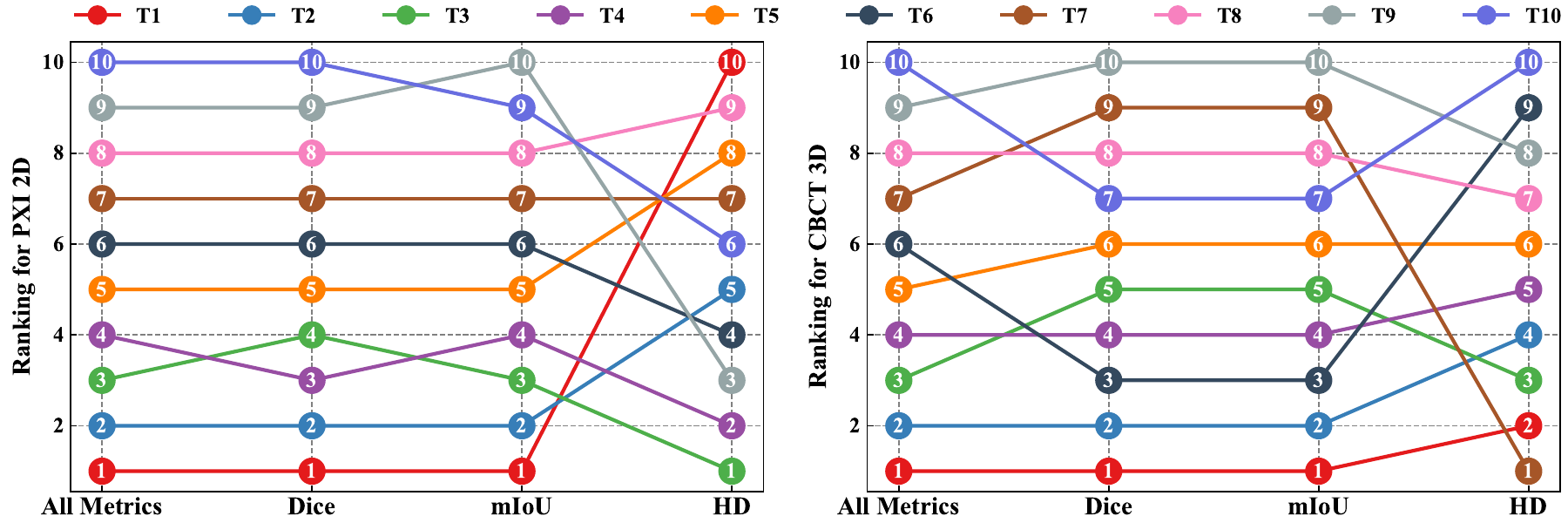}
\caption{Ranking stability analysis for the top 10 teams with individual and weighted metrics. (a) 2D tooth segmentation track, (b) 3DCBCT tooth segmentation track.
"All Metric" is the weighted ranking of three metrics (Dice, mIoU and HD).}
\label{fig:top10_ranking_analysis}
\end{figure*}

\begin{figure*}[!t]
\centering

\begin{subfigure}{0.49\textwidth}
\centering
\includegraphics[width=\textwidth]{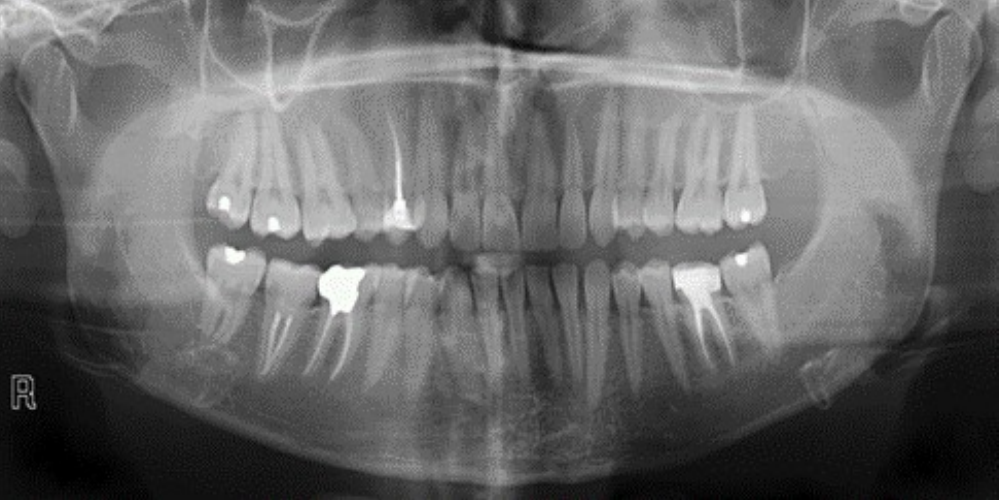}
\caption{Adult Tooth}
\label{fig:images_a}
\end{subfigure}
\hfill
\begin{subfigure}{0.49\textwidth}
\centering
\includegraphics[width=\textwidth]{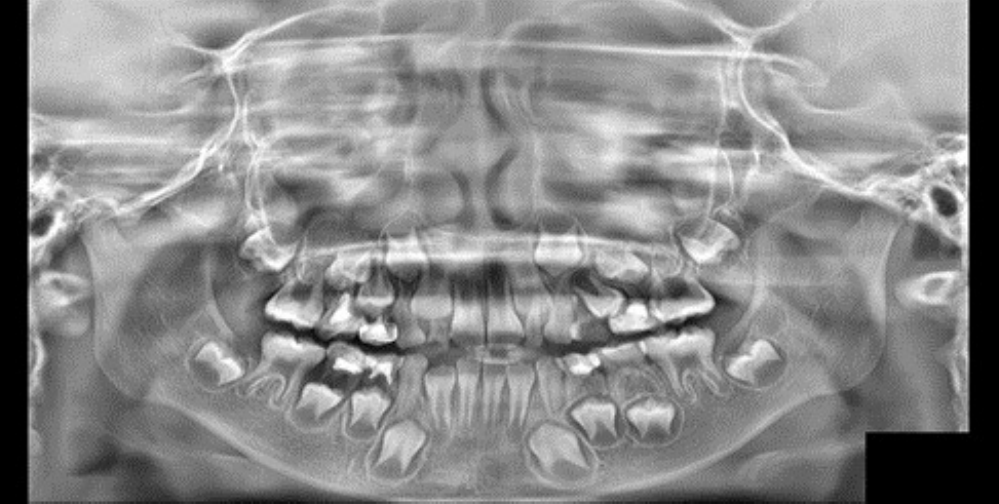}
\caption{Children Tooth}
\label{fig:images_b}
\end{subfigure}

\bigskip

\begin{subfigure}{0.49\textwidth}
\centering
\includegraphics[width=\textwidth]{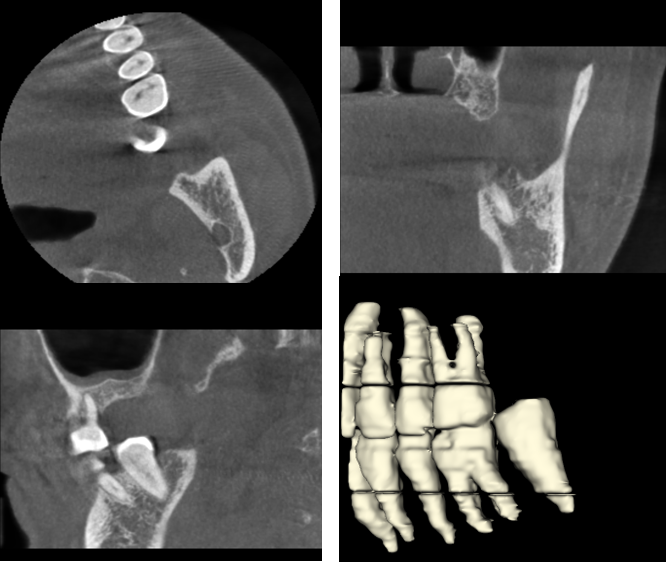}
\caption{CTooth}
\label{fig:images_c}
\end{subfigure}
\hfill
\begin{subfigure}{0.49\textwidth}
\centering
\includegraphics[width=\textwidth]{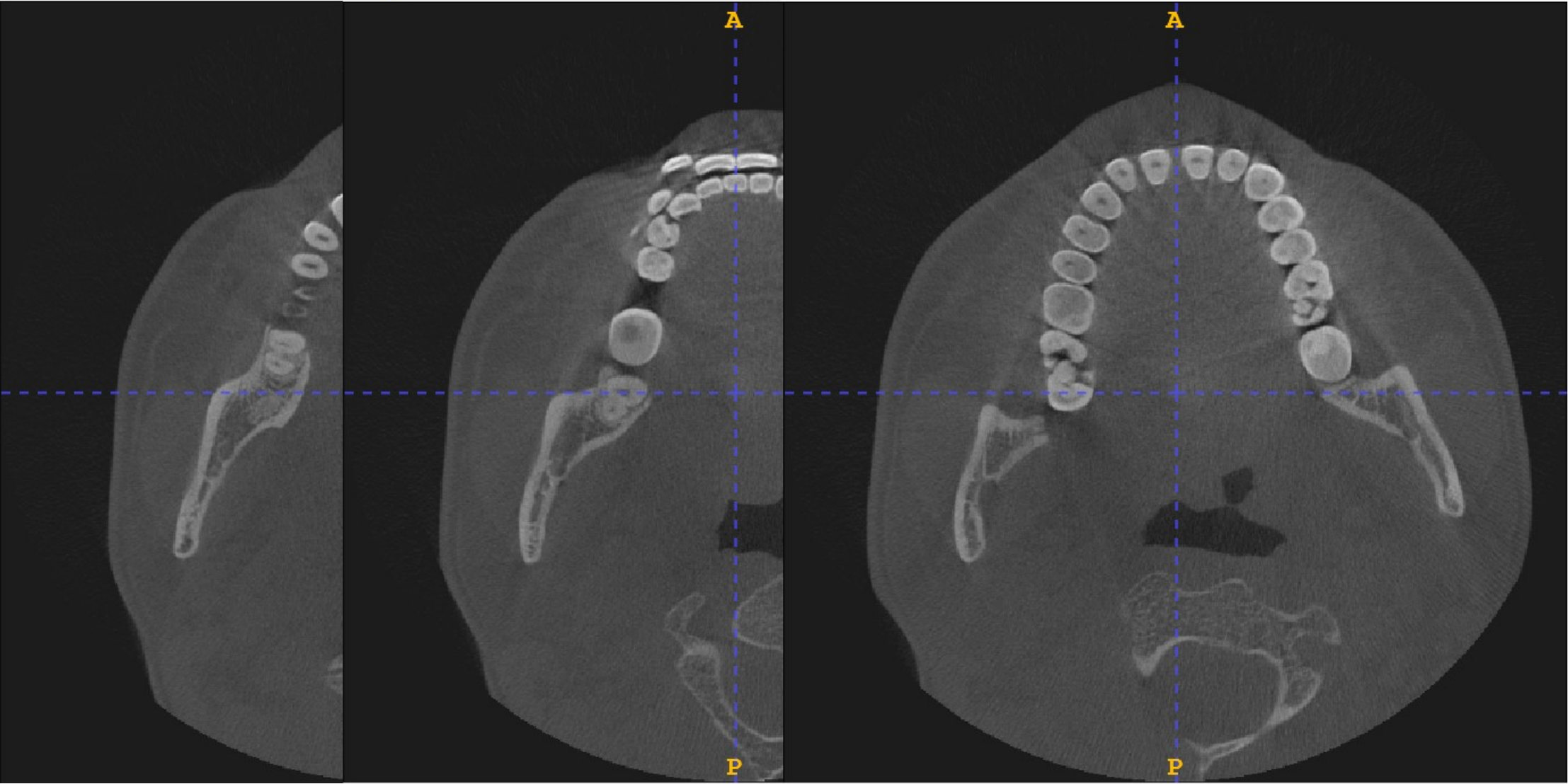}
\caption{STS-3DTooth}
\label{fig:images_d}
\end{subfigure}

\caption{The illustration of examples in the dataset.}
\label{fig:images}
\end{figure*}
\textbf{Rank 5:} Chen et al. proposed an improved network and pseudo-label generation method based on nnUNet \cite{X-7}, an axial attention mechanism \cite{X-40}, and a positional correction module.
The axial attention mechanism decomposed the 3D space into multiple independent dimensions and calculated the attention weights for each dimension separately.
It then combined them to better handle irregular shapes and localized structures, improving the ability to capture tiny structures by focusing on both horizontal and vertical structures in the image and thus enhances the accuracy of segmentation.
The position correction module was designed to accurately mark the boundaries of the teeth and exclude segmentation errors.
This module first determined the width range of the tooth.
Then, it performed the following four key steps: determining the upper left and lower right corner points of the tooth, calculating an exact position point to correct the tooth position, constructing a region box that accurately defines the extent of the tooth, and finally filtering the data to ensure that only the true tooth structure data is retained.
This process eliminated the possibility of segmentation errors.
The method leveraged unlabelled data through 5-fold cross-validation and high-quality pseudo-label generation. Subsequently, the pseudo-labels were continuously optimized through iterative training, screening for stable pseudo-labels,
and finally used the nnUNet model with axial attention to ensure that the model could perform well in the final tooth structure segmentation task.

\textbf{Rank 7:} Li et al. proposed a weak mutual consistency network (Weak MC-Net) and an entropy-based mean-teacher network \cite{X-21}.
The overall framework of the weak MC-Net referenced MC-Net+ \cite{X-20} with modifications based on VNet \cite{X-41}.
The weak MC-Net focused on the last layer in the decoder, using different up-sampling methods to better learn the features of different regions of the tooth.
In order to better distinguish between teeth and background at the macro level,
Li et al. used the CBAM module \cite{X-42} instead of the first skip connection operation.
Then, they calculated the entropy of the student model prediction results based on Mixed Transformer \cite{X-5} and used it as the weight of the consistency loss function to guide the network to learn towards targets with low entropy \cite{X-22}.
Since the outputs obtained from the two different up-sampling methods of Weak MC-Net were different perspectives on tooth learning and thus had discrepancies, calculating the MSE loss of the two outputs of the model from each other can help the outputs of the two branches converge and thus help the model better learn tooth information from unlabelled data.

\section{Evaluation Results and Ranking Analysis}
Table \ref{tab:2d3dperform} shows the DSC, mIoU, and HD for the 2D PXI semi-supervised segmentation and the 3D CBCT image tooth semi-supervised segmentation tasks. In the next subsections, we analyze the DSC, mIoU, and HD for both tasks in the STS Challenge via box plots, as well as the statistical analysis of significance. The statistical analyses were performed using a one-sided Wilcoxon signed rank test with a significance level of 5\%, which is used in many analyses of challenge results. For concise and unambiguous characterization, we focus on the top 10 teams.

Fig. \ref{fig:result2d} and Fig. \ref{fig:result3d} show box plots and statistical significance plots for the Dice, mIoU, and HD indicators of the top 10 teams for 2D and 3D tooth segmentation. The results for Dice, mIoU, and HD are presented in subfigures (a) and (d), (b) and (e), and (c) and (f), respectively. The teams are ranked from left to right in each subfigure. The statistical maps show the statistically significant differences between diverse metrics for the top 10 teams (p $<$ 0.05). Orange indicates that the performance of teams on the X-axis is significantly different from that of the teams on the Y-axis, while blue indicates the opposite.

\subsection{Dice metric analysis of Task1}
The Dice metric of all the top 10 teams is above 92\%, as demonstrated in Table \ref{tab:2d3dperform} and Fig. \ref{fig:result2d} (a). The T1 achieves the highest Dice score of 93.92 $\pm$ 4.99, while the T10 obtains the lowest score of 92.99 $\pm$ 4.05. Table \ref{tab:2d3dperform} reveals that T1, T2, and T4 rank among the top three in terms of Dice performance. Considering the substantial weight assigned to the Dice metric during final scoring, teams excelling in this evaluation metric also tend to achieve higher rankings, and vice versa.

Statistical significance analysis (Fig. \ref{fig:result2d} (d)) demonstrates significant differences between the Dice scores of the top three teams (T1, T2, and T4). Although there is no notable distinction between the highest score obtained by T1 and the Dice score achieved by T3, both teams outperform all others significantly. Teams belonging to the second-ranked (T2) and third-ranked (T4) positions exhibit considerably higher Dice scores compared to other participating teams. These results underscore the necessity of employing a Dice evaluation metric for tooth segmentation assessment.

\subsection{mIoU metric analysis of Task1}
The Dice metric only analyzes the foreground region of the segmentation result. In contrast, mIoU evaluates both the foreground and background of the segmentation result. Table \ref{tab:2d3dperform} shows that the ranking trend of the team's score in mIoU is similar to that of Dice. Most teams achieve high Dice scores and, correspondingly, better performance on the mIoU metric. However, the mIoU value is higher and the variance is lower, as shown in Fig. \ref{fig:result2d} (b). This may be because the mIoU indicator considers background segmentation, and the background occupies a relatively large proportion of the image, resulting in a relatively high average mIoU. The top three teams in the mIoU score were T1, T2, and T3, with the highest score of 98.30±0.94 achieved by the T1 team. As shown in Fig. \ref{fig:result2d} (e), there is also a significant difference between teams with higher scores. Therefore, when evaluating the performance of a segmentation model, it is necessary to consider background segmentation and introduce the mIoU metric to evaluate the algorithm.

Moreover, there was no statistical difference between the highest-scoring mIoU team T1 and the third-ranked team T3, nor between the second-ranked team T2 and the third-ranked team T3. It is worth noting that although the absolute difference in mIoU scores is small (less than 0.5\%), most of the top-ranked teams have significantly improved compared to the bottom-ranked teams. These results demonstrate that both the Dice and mIoU metrics are suitable for evaluating segmentation results simultaneously. It is reasonable to assign more weight to the Dice metric.

\begin{table*}[!t]
\centering
\caption{Summary of the benchmark methods of top seven teams for 2D tooth segmentation.}
\label{tab:top10_PXI_summary}
\setlength{\tabcolsep}{4.4mm}      
\renewcommand\arraystretch{1.3}      
\begin{tabular}{lcccl} 
\hline
\toprule
Rank                & Framework                    & Network                           & Semi-supervised Strategy                   & Highlight Methods                                                                          \\ 
\hline
\multirow{4}{*}{1}  & \multirow{4}{*}{One-stage}   & \multirow{4}{*}{SegResNet, U-Net} & \multirow{4}{*}{Consistency learning}      & Slide window strategy;                                                                     \\ 
                    &                              &                                   &                                            & MSE loss for traing student model;                                                         \\ 
                    &                              &                                   &                                            & Models ensemble;                                                                           \\ 
                    &                              &                                   &                                            & Test-time augmentation;                                                                    \\ 
\hline
\multirow{5}{*}{2}  & \multirow{5}{*}{One-stage}   & \multirow{5}{*}{SwinUNTER}        & \multirow{5}{*}{--------------------}                         & Deep supervision;                                                                          \\ 
                    &                              &                                   &                                            & Suitable data augmentations;                                \\ 
                    &                              &                                   &                                            & Models ensemble;                                                                           \\ 
                    &                              &                                   &                                            & Test-teme augmentation;                                                                    \\ 
                    &                              &                                   &                                            & Removing edge  segmentation label;                                                         \\ 
\hline
\multirow{3}{*}{3}  & \multirow{3}{*}{One-stage}   & \multirow{3}{*}{U-Net, SAM}       & \multirow{3}{*}{Pseudo-label generation  } & Models ensemble;                                           \\ 
                    &                              &                                   &                                            & Hard example mining strategy;                                                              \\ 
                    &                              &                                   &                                            & Test-time augmentation;                                     \\ 
\hline
\multirow{3}{*}{4}  & \multirow{3}{*}{One-stage}   & \multirow{3}{*}{SegFormer}        & \multirow{3}{*}{Pseudo-label generation}   & Suitable data augmentations;                                \\ 
                    &                              &                                   &                                            & Test-time augmentation;                                     \\ 
                    &                              &                                   &                                            & Choosing high-quality pseudo-labels;                                                       \\ 
\hline
\multirow{3}{*}{5}  & \multirow{3}{*}{One-stage}  & \multirow{3}{*}{UperNet}          & \multirow{3}{*}{--------------------}                         & Deep supervision;                                            \\ 
                    &                              &                                   &                                            & Hard example mining strategy;                                \\ 
                    &                              &                                   &                                            & Test-time augmentation.                                      \\ 
\hline
\multirow{3}{*}{7}  & \multirow{3}{*}{Two-stage}   & \multirow{3}{*}{U-Net}            & \multirow{3}{*}{Pseudo-label generation}   & Hard example mining strategy;                                \\ 
                    &                              &                                   &                                            & A trained U-Net denoises label result;                                                     \\ 
                    &                              &                                   &                                            & Test-time augmentation;                                      \\ 
\hline
\multirow{2}{*}{8}  & \multirow{2}{*}{One-stage}   & \multirow{2}{*}{FCBformer}        & \multirow{2}{*}{--------------------}                         & Parameter averaging;                                         \\ 
                    &                              &                                   &                                            & Lion optimizer;                                                                           \\ 
\hline
\multirow{3}{*}{11} & \multirow{3}{*}{One-stage  } & \multirow{3}{*}{DecoupledSegNet}  & \multirow{3}{*}{Pseudo-label generation}   & Customized data~augmentation;  \\ 
                    &                              &                                   &                                            & Deep supervision;                                           \\ 
                    &                              &                                   &                                            & Dynamic mutual loss function;  \\
\hline
\toprule
\end{tabular}
\end{table*}

\begin{table*}[!t]
\centering
\caption{Summary of the benchmark methods of top seven teams for 3D tooth segmentation.}
\label{tab:top10_CBCT_summary}
\setlength{\tabcolsep}{3.5mm}      
\renewcommand\arraystretch{1.3}    
\begin{tabular}{lcccl} 
\hline
\toprule
Rank               & Framework                                                                              & Network                                              & Semi-supervised Strategy                                                                                & Highlight Methods                                                                                                                                                                                               \\ 
\hline
\multirow{3}{*}{1} & \multirow{3}{*}{Two-stage}                                                             & \multirow{3}{*}{nnUNet}                              & \multirow{3}{*}{Pseudo-label generation}                                                                & Maxilla and Mandible extracted by trained network;                                                                                                                                                           \\ 
                   &                                                                                        &                                                      &                                                                                                         & Label noise tackling;                                                                                                                                                                                         \\ 
                   &                                                                                        &                                                      &                                                                                                         & Suitable data augmentation;                                                                                                                                                                                    \\ 
\hline
\multirow{3}{*}{2} & \multirow{3}{*}{One-stage}                                                             & \multirow{3}{*}{nnUNet}                              & \multirow{3}{*}{Pseudo-label generation}                                                                & Customized nnUNet configure;                                                                                                                        \\ 
                   &                                                                                        &                                                      &                                                                                                         & Label noise tackling;                                                                                                                                                                                       \\ 
                   &                                                                                        &                                                      &                                                                                                         & Selecting reliable pseudo labels;                                                                                                                                                                               \\ 
\hline
\multirow{3}{*}{3} & \multirow{3}{*}{Two-stage}                                                             & \multirow{3}{*}{U-Net}                               & \multirow{3}{*}{\begin{tabular}[c]{@{}l@{}}Pseudo-label generation \\ and consistency learning\end{tabular}} & Adaptive confidence level;                                                                                                                                                                                   \\ 
                   &                                                                                        &                                                      &                                                                                                         & Split Coordinate Attention;                                                                                                                                                                                  \\ 
                   &                                                                                        &                                                      &                                                                                                         & Frequency transform data augmentation;                                                                                \\ 
\hline
\multirow{3}{*}{4} & \multirow{3}{*}{Two-stage} & \multirow{3}{*}{U-Net 3D}                            & \multirow{3}{*}{\begin{tabular}[c]{@{}l@{}}Pseudo-label generation\\ and consistency learning\end{tabular}}  & Anti-noise segmentation framework;                                                                                                                 \\ 
                   &                                                                                        &                                                      &                                                                                                         & Test-time augmentation;                                                                                                                                                          \\ 
                   &                                                                                        &                                                      &                                                                                                         & Designed post-processing operation;                                                                                                                                              \\ 
\hline
\multirow{2}{*}{5} & \multirow{2}{*}{One-stage}                                                             & \multirow{2}{*}{nnUNet}                              & \multirow{2}{*}{Pseudo-label generation}                                                                & Attention mechanism;                                                                                                 \\ 
                   &                                                                                        &                                                      &                                                                                                         & Position correction post-processing;                                                                                                                \\ 
\hline
\multirow{3}{*}{6} & \multirow{3}{*}{One-stage}                                                             & \multirow{3}{*}{U-Net}                               & \multirow{3}{*}{--------------------}                                                                                      & Deep supervision;                                                                                                                                                                \\ 
                   &                                                                                        &                                                      &                                                                                                         & Suitable data augmentation;                                                                                                                                                                                 \\ 
                   &                                                                                        &                                                      &                                                                                                         & Test-time augmentation;                                                                                                                                                           \\ 
\hline
\multirow{3}{*}{7} & \multirow{3}{*}{One-stage}                                                             & \multirow{3}{*}{V-Net} & \multirow{3}{*}{Consistency learning}                                                                   & Sliding window strategy;                                                                                                                                                          \\ 
                   &                                                                                        &                                                      &                                                                                                         & Soft shapen pseudo method; \\ 
                   &                                                                                        &                                                      &                                                                                                         & Designed post-processing operation;                                                                                                                                            \\
\hline
\toprule
\end{tabular}
\end{table*}

\begin{table*}[!t]
\centering
\caption{Summary of the experimental setup of top seven teams for 2D tooth segmentation. Abbreviations: a) Data-Augmentation: Slidingwindow (S), RandomRotate (R), RandomFlip (RF), CutMix (CM), Mixup (M), ColorJitter (C), ElasticTransform (E), GaussianBlur (GB), GaussianNoise (GN), IntensityTransformations (I), RandomCropping (RC), CoarseDropout (CD), ShiftScaleRotate (SR), Random Mosaic (RM), Grid Distortion (GD), Random Occlusion (RO), PhotoMetricDistortion (PM), Remove (RE), Scaling (SC); b) Pre-processing: Normalization (N), Duplicate Image Detection (DI), Multi-Label Sample Filtering (MS), Image Rotation Augmentation (IR), Resize (RS), Cropping (CR), Padding (PA); c) Post-processing: voting averaging (V), Probability Threshold Filtering (PF), Filling Holes (F), Bilinear Interpolation (BI), Masking (M); Loss:Dice Loss (D), Lovasz Loss (L), Class Balanced Loss (CB), DynamicMutual Loss (DM); d) Optimizer:AdamW (AW).}
\label{tab:2daug}
\setlength{\tabcolsep}{1mm}        
\renewcommand\arraystretch{2.5} 
\scalebox{0.7}{
\begin{tabular}{lllllllllllllllllllllllllllllllllllllllll}
\hline
\toprule
\multirow{2}{*}{Rank} & \multicolumn{19}{c}{Data Augmentation}  & \multicolumn{7}{c}{Pre-processing}    & \multicolumn{5}{c}{Post-processing}  & \multicolumn{9}{c}{Loss}     \\ 
\cmidrule(r){2-20} \cmidrule(r){21-27} \cmidrule(r){28-32} \cmidrule(r){33-41} 
& S & RF & R & CM & M & C & E & GB & GN & I & CD & SR & RC & RM & GD & RO & PM & RE & SC & N & DI & MS & IR & RS & CR & PA & V & PF & F & BI & M & BCE & D & L & CB & IoU & HD & DM & MSE & CE                                                               \\ \hline
1  & $\checkmark$                               & $\checkmark$                                &                                 &                                  &                                 &                                 &                                 & $\checkmark$                                & $\checkmark$                                & $\checkmark$                               &                                  & $\checkmark$                                &                                  &                                  &                                  &                                  &                                  &                                  &                                  & $\checkmark$                               &                                  &                                  &                                  &                                  &                                  &                                  &                                 &                                  &                                 &                                  &                                 & $\checkmark$                                 &                                 &                                 &                                  &                                   &                                  &                                  & $\checkmark$                                 &                            \\ 
2 &                                 & $\checkmark$                                &                                 &                                  &                                 &                                 &                                 & $\checkmark$                                & $\checkmark$                                & $\checkmark$                               &                                  & $\checkmark$                                &                                  &                                  &                                  &                                  &                                  &                                  &                                  & $\checkmark$                               &                                  &                                  &                                  &                                  &                                  &                                  &                                 &                                  &                                 &                                  & $\checkmark$                               & $\checkmark$                                 & $\checkmark$                               &                                 &                                  &                                   &                                  &                                  &                                   &                                \\ 
3  &                                 & $\checkmark$                                & $\checkmark$                               &                                  & $\checkmark$                               &                                 &                                 &                                  &                                  &                                 & $\checkmark$                                &                                  &                                  &                                  &                                  &                                  &                                  &                                  &                                  &                                 & $\checkmark$                                & $\checkmark$                                &                                  &                                  &                                  &                                  &                                 &                                  &                                 & $\checkmark$                                &                                 &                                   & $\checkmark$                               & $\checkmark$                               &                                  &                                   &                                  &                                  &                                   & $\checkmark$             \\ 
4         &                                 & $\checkmark$                                &                                 &                                  &                                 & $\checkmark$                               & $\checkmark$                               & $\checkmark$                                &                                  & $\checkmark$                               &                                  & $\checkmark$                                &                                  &                                  &                                  &                                  &                                  &                                  &                                  & $\checkmark$                               &                                  &                                  &                                  &                                  &                                  &                                  &                                 &                                  & $\checkmark$                               &                                  &                                 & $\checkmark$                                 & $\checkmark$                               &                                 &                                  &                                   &                                  &                                  &                                   &     \\ 
5                   &                                 & $\checkmark$                                &                                 &                                  &                                 &                                 &                                 &                                  &                                  &                                 &                                  & $\checkmark$                                & $\checkmark$                                & $\checkmark$                                &                                  &                                  & $\checkmark$                                &                                  &                                  &                                 &                                  &                                  & $\checkmark$                                &                                  &                                  &                                  &                                 & $\checkmark$                                &                                 &                                  &                                 &                                   &                                 &                                 & $\checkmark$                                &                                   &                                  &                                  &                                   &                              \\ 
6       &                                 & $\checkmark$                                &                                 &                                  &                                 &                                 &                                 &                                  & $\checkmark$                                & $\checkmark$                               &                                  & $\checkmark$                                & $\checkmark$                                &                                  & $\checkmark$                                & $\checkmark$                                &                                  &                                  &                                  & $\checkmark$                               &                                  &                                  &                                  & $\checkmark$                                &                                  &                                  & $\checkmark$                               &                                  &                                 &                                  &                                 & $\checkmark$                                 & $\checkmark$                               &                                 &                                  & $\checkmark$                                 & $\checkmark$                                &                                  &                                   &                       \\ 
7               &                                 & $\checkmark$                                & $\checkmark$                               &                                  & $\checkmark$                               & $\checkmark$                               & $\checkmark$                               & $\checkmark$                                & $\checkmark$                                & $\checkmark$                               &                                  &                                  &                                  &                                  &                                  &                                  &                                  &                                  &                                  & $\checkmark$                               &                                  &                                  &                                  &                                  &                                  &                                  &                                 &                                  &                                 &                                  &                                 & $\checkmark$                                 & $\checkmark$                               &                                 &                                  &                                   &                                  &                                  &                                   &                            \\ 
8                         &                                 & $\checkmark$                                &                                 &                                  &                                 & $\checkmark$                               &                                 & $\checkmark$                                &                                  &                                 &                                  & $\checkmark$                                &                                  &                                  &                                  &                                  &                                  &                                  &                                  & $\checkmark$                               &                                  &                                  &                                  &                                  & $\checkmark$                                & $\checkmark$                                &                                 &                                  &                                 &                                  &                                 &                                   & $\checkmark$                               &                                 &                                  &                                   &                                  &                                  &                                   & $\checkmark$             \\ 
11                     &                                 & $\checkmark$                                &                                 &                                  &                                 &                                 &                                 &                                  & $\checkmark$                                & $\checkmark$                               &                                  &                                  &                                  &                                  &                                  &                                  &                                  & $\checkmark$                                & $\checkmark$                                &                                 &                                  &                                  &                                  &                                  &                                  &                                  &                                 &                                  &                                 &                                  &                                 &                                   &                                 &                                 &                                  &                                   &                                  & $\checkmark$                                &                                   &                               \\ 
\hline
\toprule
\end{tabular}
}

\end{table*}

\begin{table*}[!t]

\centering
\caption{Summary of the experimental setup of top seven teams for 3D tooth segmentation. Abbreviations: a) Data Augmentation: FTA Module (F), RandomRotate (R), RandomFlip (RF), CutMix (CM), Mixup (M), ColorJitter (C), ElasticTransform (E), GaussianBlur (GB), Gaussian Noise (GN), Intensity Transformations (I), Random Cropping (RC); b) Pre-processing: boundary smoothing (BS), Normalization (N), Thresholding (T), Resampling (R), Random Cropping (RC); c) Post-processing: boundary smoothing (BS), Connected Component Analysis (CC), Morphological Operations (MO), Filling Holes (F); Loss: Dice (D), Focal (F). }
\label{tab:3daug}
\setlength{\tabcolsep}{1.6mm}        
\renewcommand\arraystretch{1.3}  
\begin{tabular}{lllllllllllllllllllllllll}
\hline
\toprule
\multirow{2}{*}{Rank} & \multicolumn{11}{c}{Data Augmentation} & \multicolumn{5}{c}{Pre-processing} & \multicolumn{4}{c}{Post-processing}                                                                     & \multicolumn{4}{c}{Loss} \\ 
\cmidrule(r){2-12} \cmidrule(r){13-17} \cmidrule(r){18-21} \cmidrule(r){22-25}
& RC & F & R & RF & CM & M & C & E & GB & GN & I & BS & N & T & R & RC & BS & CC & MO & F & D & CE & F & MSE       
\\ \hline
1     & $\checkmark$                                 &                                 & $\checkmark$                               & $\checkmark$                                &                                  &                                 & $\checkmark$                               & $\checkmark$                               &                                  & $\checkmark$                                & $\checkmark$          & $\checkmark$                                &                                 &                                 &                                 &             & $\checkmark$                                &                                 &                                 &            & $\checkmark$                               & $\checkmark$                                &                                 &              \\ 
2                                          & $\checkmark$                                &                                 & $\checkmark$                               & $\checkmark$                                &                                  &                                 & $\checkmark$                               & $\checkmark$                               &                                  & $\checkmark$                                & $\checkmark$          &                                  & $\checkmark$                               &                                 & $\checkmark$                               & $\checkmark$           &                                  & $\checkmark$                               & $\checkmark$                               &            & $\checkmark$                               & $\checkmark$                                &                                 &              \\ 
3                                         &                                  &                                 & $\checkmark$                               & $\checkmark$                                & $\checkmark$                                &                                 & $\checkmark$                               &                                 & $\checkmark$                                &                                  &            &                                  & $\checkmark$                               & $\checkmark$                               &                                 &             &                                  &                                 &                                 &            & $\checkmark$                               & $\checkmark$                                &                                 &   \\ 
4                                           &                                  &                                 & $\checkmark$                               & $\checkmark$                                &                                  &                                 &                                 &                                 &                                  &                                  &            &                                  & $\checkmark$                               &                                 &                                 &             &                                  & $\checkmark$                               &                                 & $\checkmark$          & $\checkmark$                               & $\checkmark$                                &                                 &              \\ 
5                                        & $\checkmark$                                &                                 & $\checkmark$                               & $\checkmark$                                &                                  &                                 & $\checkmark$                               & $\checkmark$                               &                                  & $\checkmark$                                & $\checkmark$          &                                  & $\checkmark$                               &                                 & $\checkmark$                               & $\checkmark$           &                                  &                                 &                                 &            & $\checkmark$                               & $\checkmark$                                &                                 &              \\ 
6                                          &                                  &                                 &                                 & $\checkmark$                                & $\checkmark$                                & $\checkmark$                               &                                 & $\checkmark$                               & $\checkmark$                                & $\checkmark$                                & $\checkmark$          &                                  &                                 &                                 & $\checkmark$                               & $\checkmark$           &                                  &                                 &                                 &            & $\checkmark$                               &                                  & $\checkmark$                               \\ 
7                                           & $\checkmark$                                &                                 &                                 & $\checkmark$                                &                                  &                                 &                                 &                                 &                                  &                                  &            &                                  & $\checkmark$                               & $\checkmark$                               & $\checkmark$                               & $\checkmark$           &                                  & $\checkmark$                               & $\checkmark$                               & $\checkmark$          &                                 &                                  &                                 & $\checkmark$            \\ 
\hline
\toprule
\end{tabular}
\end{table*}

\subsection{HD metric analysis of Task1}
\begin{figure*}[!t]
\centering
\includegraphics[width=\textwidth]{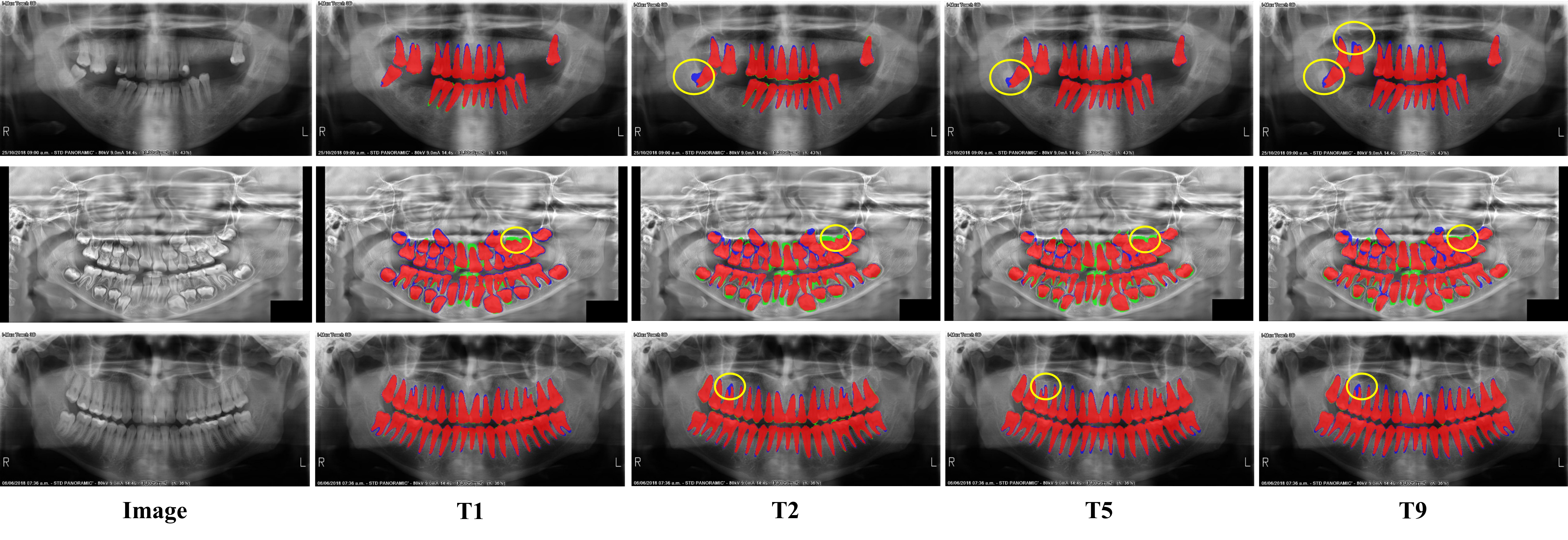}
\caption{Visualization of 2D PXI segmentation results representing the team and segmentation results for different tooth types. Left to right: original images and results form teams T1, T2, T5, T9. The first row depicts cases of tooth loss, the second row illustrates a combination of caries and permanent teeth, and the third row represents normal dental conditions. The red color indicates the region where the ground truth and prediction results coincide, blue represents the under-segmentation area, and green corresponds to the over-segmentation area.}
\label{fig:prediction2d}
\end{figure*}

The normalized HD metric is utilized to assess the edge effect of the segmentation outcomes. Table \ref{tab:2d3dperform} displays that the T3, T4, and T9 teams have the top three HD scores, respectively, and their scores have low dispersion. However, it is worth noting that the teams with the highest Dice and mIoU scores do not always have the highest HD scores. For instance, the team ranked third in the HD metric, T9, and achieved a score of 0.0234 $\pm$ 0.0159. However, its Dice and mIoU scores were not satisfactory. As a result, T9 has fewer segmentation errors in the tooth region but relatively more edge-fitting errors in the boundary segmentation, as shown in Fig. \ref{fig:prediction2d}.
Notably, even if the difference in HD scores is not significant, it can still affect the ranking when the Dice and mIoU scores are similar. Dice, mIoU, and HD are complementary metrics that should be used together to comprehensively evaluate the performance of segmentation algorithms.

Table \ref{tab:dataset2d} (f) shows that T3 with the highest HD score is significantly better than that of most teams, and the score has been significantly improved. However, most of the other teams did not make significant improvements. These results demonstrate that the segmentation evaluation method that considers the edge effect is more robust and reliable.


\subsection{Dice metric analysis of Task2}
Table \ref{tab:2d3dperform} demonstrates that T1, T2, and T6 achieved the highest Dice scores for the tooth 3D segmentation task. The top 10 teams achieved Dice scores ranging from 76.64 $\pm$ 6.47 to 84.42 $\pm$ 5.41, with a significant gap between the highest and lowest scores. The shortage of labelled training data for 3D tasks may be the reason for this. There are only a limited number of cases available. Furthermore, semi-supervised segmentation is more challenging, resulting in significant differences in optimization results among different teams. Table \ref{tab:2d3dperform} shows that similar results can be achieved with 2D tooth segmentation; teams with higher Dice scores are ranked higher in the final weighted scores.

Based on the statistical significance analysis as shown in Figure \ref{fig:result3d} (d), it is evident that there is no significant difference between T1 and T2 with Dice scores. However, there is a statistical difference between T1 and T6. The results indicate that the teams at the top of the rankings are significantly better than those at the bottom. This demonstrates that the Dice metric is an effective evaluation tool for 3D tooth segmentation and can accurately assess the quality of segmentation results.

\subsection{mIoU metric analysis of Task2}
Table \ref{tab:2d3dperform} shows that the trend of the area-based mIoU metric is consistent with that of the Dice metric score. The top three teams are T1, T2, and T6, respectively, but there is little difference in mIoU scores among the top ten teams, where the highest mIoU score is 86.61±3.84, and the lowest is 81.11±3.96. Additionally, when combined with Fig. \ref{fig:result3d} (b), it is evident that there are fewer outliers in mIoU scores across different cases. Background segmentation is considered in the mIoU metric, which explains why the tooth foreground area in 3D CBCT, being more sparse, does not significantly affect the mIoU score.

Fig. \ref{fig:result3d} (e) shows that the top three teams have significantly higher mIoU scores than most other teams. There was no significant difference between the highest score of T1 and the score of T2, both of which outperformed all other teams. The third-placed team, T6, scored significantly higher than all the other teams in terms of mIoU. The visualization of segmentation results (Fig. \ref{fig:prediction3d}) shows that the overall weighted ranking T6 team had fewer segmentation errors than T10. Therefore, it is reasonable to adopt the mIoU metric to evaluate segmentation results, as it considers the background and can effectively assess both over-segmentation and under-segmentation.

\subsection{HD metric analysis of Task2}
\begin{figure*}[!t]
\centering
\includegraphics[width=\textwidth]{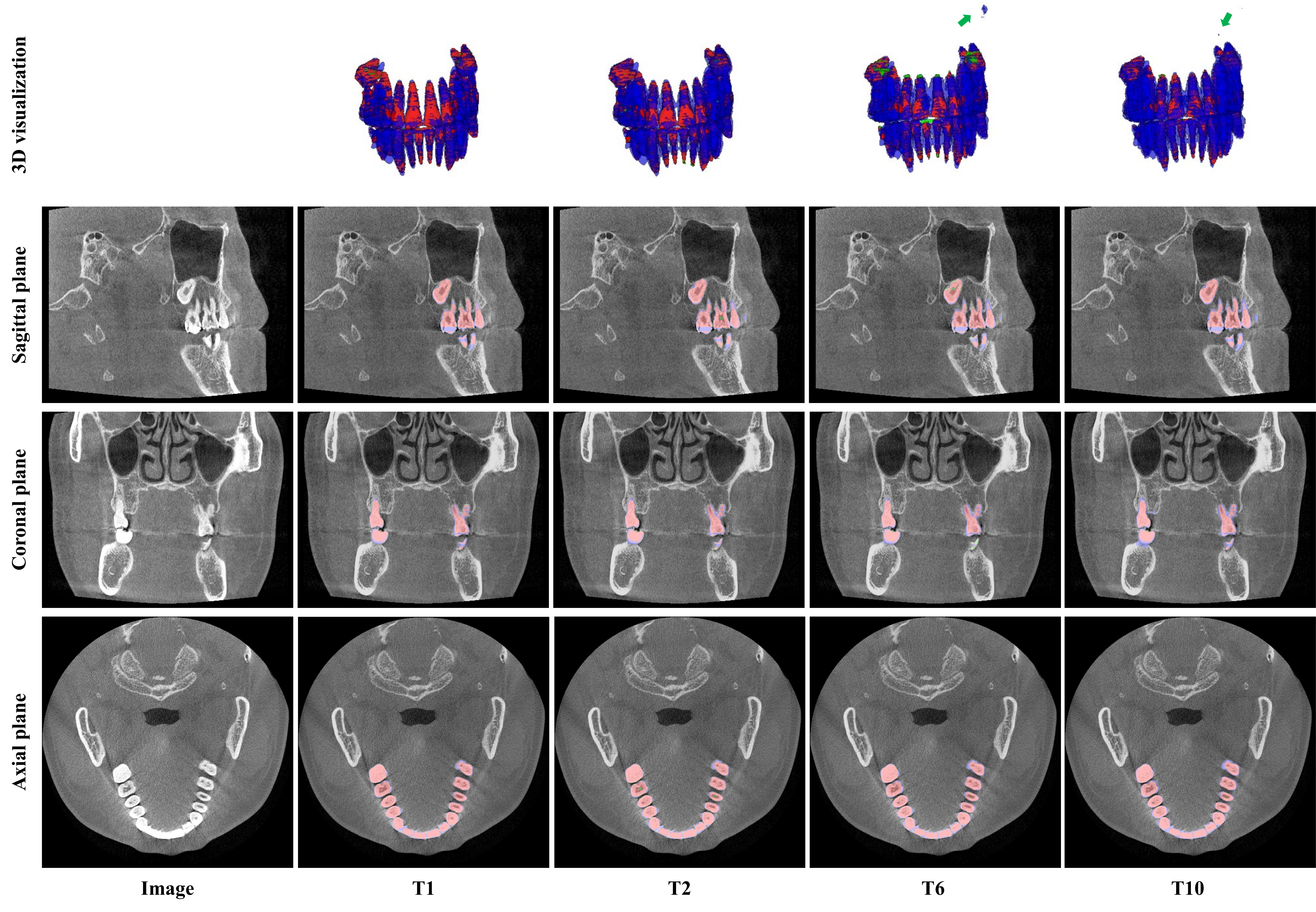}
\caption{3D visualization of CBCT segmentation results (the first row), and three-view segmentation results (the last three rows) for representative teams. Left to right: original images and segmentation results from
teams T1, T2, T6, and T10. The red color indicates the region where the ground truth and prediction results coincide, blue represents the under-segmentation area, and green corresponds to the over-segmentation area.}
\label{fig:prediction3d}
\end{figure*}

Table \ref{tab:2d3dperform} shows that the normalized HD scores of the top ten are mostly close and relatively stable, with no outliers. The team with the lowest HD error was T7, with a score of 0.1580 $\pm$ 0.1080. It is worth noting that only two teams in the top 10, T6 and T10, have HD errors greater than 0.2. This causes the top three T6 scores in the Dice and mIoU indicators to be pulled down in the ranking due to the HD scores in the evaluation of the split edge. Fig. \ref{fig:prediction3d} illustrates that there are some discrete label regions in the predicted segmentation results of both teams, resulting in low HD scores for the predicted results.

Fig. \ref{fig:result3d} (f) shows that with the exception of T6 and T10 teams, there is no significant difference in HD scores between the remaining teams, but they are all significantly better than T6 and T10. The results demonstrate that the HD evaluation metric is a crucial indicator for evaluating segmentation methods. Its inclusion can enhance the comprehensiveness of the evaluation of segmentation results.

\section{Discussion}
\subsection{Organization}
The STS Challenge provides services for both tracks on the Alibaba Tianchi platform. The organizers employ an online evaluation method to prevent cheating without disclosing the test set labels. Consequently, all participants must submit their predictions to the Alibaba Tianchi system. The Alibaba Tianchi system automatically calculates evaluation metrics against the ground truth to determine participants' scores. In the end, 434 teams participated in the preliminary stage across both tracks, with 64 teams advancing to the finals. Notably, there were a higher number of teams participating in the 2D task.

\subsection{Ranking stability analysis}
In order to more accurately evaluate the algorithmic strengths of different participating teams, the three evaluation metrics (Dice, mIoU, and HD) were assigned varying weights in the final ranking. Specifically, a weight of 40\% was assigned to the Dice metric, while the mIoU metric, which accounts for both background and foreground, received a weight of 30\%, and the final HD metric score was set at 30\%. To assess how individual metrics influence the final rankings, the ranking stability of the top 10 teams for the tasks of 2D panoramic tooth segmentation and 3D CBCT tooth segmentation is illustrated in Fig. \ref{fig:top10_ranking_analysis}. There are noticeable fluctuations in their HD metrics for both 2D and 3D tasks.

These variations could be attributed to differing priorities among teams. For instance, the top-ranking team in the 2D task focused on the Dice and mIoU metrics, achieving first place in both, yet had the poorest HD score among the top 10. Similarly, in the 3D task, the team ranked 7th overall was 9th in the Dice and mIoU metrics but secured the 1st position in the HD metrics.

Moreover, based on the results in Fig. \ref{fig:prediction3d}, it appears that this situation may be due to the presence of discrete outliers in the segmentation results. Such outliers can typically be addressed through post-processing methods that preserve the maximum connectivity domain. For instance, Li, ranked 7th, initially conducted hyperparameter selection for subsequent connectivity domain analysis and morphological operations based on a priori knowledge. Then, for the prediction result, Li used connectivity domain analysis and morphological operations, including erosion, expansion, and filling of voids, to handle small pixel points outside the main region of the tooth, ultimately obtaining the processed prediction result. This post-processing scheme effectively addresses the discrete points outside the main region of the tooth produced by the segmentation network, significantly improving the HD metrics.

Therefore, a more comprehensive consideration of various metrics to evaluate the stability and consistency of segmentation algorithms is a suitable direction for future algorithm evaluation development.

\subsection{Analysis of the top-ranked methods on 2D Task}
Table \ref{tab:2daug} illustrates that certain participants utilized a network architecture approach employing two segmentation networks for segmenting the same image, yielding commendable results.
The first-place participant, Zhuang et. al, employed both SegResNet\_64 and DynUNet segmentation networks, leveraging the powerful feature extraction capabilities of SegResNet\_64 and the dynamic convolution kernel adjustment ability of DynUNet to aid in tooth segmentation. This approach does improve performance when processing large numbers of oral x-ray images by fusing multiple teacher and student models into a single robust model. However, this complex model structure may lead to computational inefficiency and greatly consume hardware resources. 

Similarly, the third-place team, Liu et al., utilized an EfficientNetv2-based UNet \cite{tan2021efficientnetv2} in conjunction with the SAM \cite{kirillov2023segment} model.
Both networks were fine-tuned after pre-training, resulting in faster and more accurate tooth area segmentation.
However, Liu et al. may have filtered out some anomalous but practically useful samples during the process in data preprocessing. In addition, SAM was trained on 11 million image datasets to obtain the pre-training weights, and whether Liu et al.'s fine-tuning of them on a much smaller dataset than the pre-training is effective in learning new knowledge deserves deeper consideration. 

Wang et al. who received the fifth ranking used large and complex models that may lead to higher computational costs and low tolerance for misclassified backgrounds, while the loss of category imbalance in the enhancement prospect and online hard sample mining may introduce noise that affects the model generalization ability. In addition, the data enhancement techniques and post-processing strategies employed, while improving performance, also increase processing time and tuning complexity, especially when setting effective thresholds to distinguish between teeth and background.

Notably, some participants combined Transformer structures with traditional CNNs, utilizing the Transformer's ability to capture global dependencies in the input data and the CNN's capability to extract local features. The second-place team, Yang et al., expanded the backbone scale and embedded sub-models into SwinUNETR\cite{X-45}, incorporating deep supervision, channel attention mechanisms, and a feature extraction module inspired by PointRend to construct their SwinUNETR\_HS network.

It is apparent from Table \ref{tab:2daug} that among the top seven teams, only one participant employed a multi-stage training approach. Specifically, the seventh-place team, Lin et al., divided the tooth segmentation task into stages of easy label prediction, challenging label regeneration, and final model training, enhancing the quality of segmentation maps while reducing pseudo-label noise. Although this multi-stage task strategy is commendable, the choice of the simplest Unet structure as the backbone network may explain why the segmentation results were not particularly outstanding. Moreover, the method proposed by Lin et al. is expected to improve the model performance with staged training and the use of denoisers, there are two possible problems with this approach: first, the model relies on high-quality pseudo-labels to guide the learning of unlabeled data, and incorrect pseudo-labels may lead to error accumulation; second, the three-stage training process increases the complexity of the training, and is very sensitive to the selection of hyper-parameters and enhancement strategies which makes it difficult to ensure that the final model presents high accuracy and robustness.

The eighth-ranked FCBformer model proposed by Song et al. Although it skillfully fuses FCNs and Transformers, as well as accelerates model convergence through pre-training to improve performance, it may suffer from two problems: firstly, the complex architecture and training strategy may lead to high computational cost and sensitivity to hyper-parameter tuning; and secondly, although pre-training the model helps to improve the accuracy, an over-reliance on pre-training the model may limit the model's ability to adapt to new data.

The eleventh-ranked Xue et al. used an automatically generated data augmentation and DecoupledSegNets model architecture. The automatically generated data augmentation may result in unrealistic samples that affect the actual efficacy of the model and the special structure makes the network potentially lack robustness in dealing with uncommon tooth morphology; furthermore, the multi-step segmentation process increases the complexity of the model and may result in performance degradation in some specific cases.

In Table \ref{tab:top10_PXI_summary}, it can be observed that only four participants employed semi-supervised learning to utilize more than half of the unlabelled data fully, yet three of the top four utilized semi-supervised algorithms. It is worth mentioning that most participants used pseudo-label generation; notably, only the first-place participant employed the classical teacher-student network framework for semi-supervised learning. It is noticeable from Table \ref{tab:2daug} that almost all top seven participants adopted a combination of various loss functions as their final loss function, indicating that using a mix of loss functions is a common strategy.

\subsection{Analysis of the top-ranked methods on 3D Task}
According to Table \ref{tab:3daug}, it can be observed that four of the top seven teams have adopted nnUnet \cite{X-7} as their backbone network architecture, remarkably including the top three. nnUNet is a powerful image segmentation framework that utilizes both 2D and 3D UNet structures. It adaptively tunes all hyperparameters, in addition to covering aspects such as data augmentation. Some of these teams also made some modifications to the nnUnet network. The third-place team, Wang et al., proposed the FTA data augmentation module, which utilizes the Fourier transform for data enhancement. Li et al. constructed a larger and deeper nnUnet by increasing the number of layers and channels in the nnUnet. In fifth place, Chen et al. added the Axial Attention Mechanism (AAM) and the Ositional Correction Module (OCM). It is clear that the above four teams using nnUNet primarily adopted the framework's data augmentation strategies, with one team proposing its data augmentation module. The third-place team, Wang et al., proposed the FTA data augmentation module, which utilizes the Fourier transform for data enhancement.

As can be seen in Table \ref{tab:3daug}, compared to the 2D track, where only one of the top seven players adopted multi-stage training and achieved less satisfactory results, in the 3D track, three of the top four teams adopted a multi-stage training strategy by dividing the tooth segmentation task into multiple stages to achieve it. Among them, the first-place Li et al. effectively improved the segmentation accuracy by first segmenting the maxilla and mandible strategies. The strategy first segments the maxilla and mandible, on the basis of which the tooth region is segmented. However, the position of the maxilla and mandible inferred first largely determines the accuracy of tooth segmentation. Therefore, its accuracy needs to be further improved. 

In third place, Wang et al. used a fully supervised approach to generate and screen out a small number of pseudo-labels in the first stage of training and then used a semi-supervised algorithm to train in the second stage. An innovative FTA data enhancement module was proposed. However, Wang et al. converted the 3D image into a 2D picture for training and feature learning, thus potentially ignoring latent features in 3D space.

In fourth place, Liu et al. applied the Mean-Teacher method for initial full-field-of-view segmentation, then strengthened the dataset with pseudo-labels for noise-resistant segmentation to improve background differentiation accuracy, and finally focused on the region of interest to achieve more detailed segmentation. Semi-supervised and anti-noise strategies are used to achieve fine-grained segmentation through three stages of training. However, the above training process is more time-consuming and the speed of the inference process is slow. In addition, too many human operations such as data selection may also affect the accuracy of its training process. Finally, as can be seen from Table \ref{tab:top10_CBCT_summary}, six of the top seven finishers used semi-supervised algorithms to make the most of those unlabelled data. 

The fifth-place winner Chen et al. designed an nnU-Net-based framework to utilize unlabeled data by modifying the second layer decoder of nnUNet and introducing 3D axial attention mechanism and pseudo-labeling mechanism. However, they only used a simple uncertainty-based pseudo-label screening method, which may generate large pseudo-label noise and thus interfere with the training process and its result accuracy.

The seventh-place winner Lin et al. proposed a Weak MC-Net, which manually identifies the ROI regions present in the dental region during the training process and crops them for training, but the above preprocessing operation leads to the inadequacy of the trained model in reasoning in the face of uncropped data.

It is noteworthy that the third-place winner, Wang et al., achieved two key milestones in their training process. In the first stage, they obtained a small number of high-quality pseudo-labels, which increased the volume of labelled data for the second stage. Then, in the second stage, they utilized an enhanced version of the Unimatch semi-supervised algorithm to effectively leverage the unlabelled data twice, fully harnessing the embedded information within it. Similarly, Liu et al. in the fourth place also utilized unlabelled data repeatedly during the multi-stage training process.

In Table \ref{tab:3daug}, we can see that the top five teams all adopted the strategy of the weighted average of dice loss and cross-entropy loss function, which is a common combination in the field of image segmentation. Dice loss can help to deal with the problem of category imbalance, but the gradient may be unstable during the training process, while cross entropy loss is more concerned about the accuracy of the prediction results and can stabilize the gradient changes during training, so the two are combined to serve as a comprehensive loss function.

\subsection{Limitation}
The STS 2023 Challenge presents two tasks to segment out teeth on the PXI and CBCT. Automatically segmenting out tooth regions can help dentists with dental diagnosis, treatment, and evaluation, and participants were asked to automatically segment tooth regions in PXI or CBCT with the help of a small amount of labelled data and a large amount of unlabelled data. Overall, for the PXI task, the top seven participants achieved a high total prediction score. This was because we created a PXI dataset containing teeth. However, the dataset still has some limitations. Firstly, patient confidentiality is a major obstacle to obtaining a valuable dataset. The data in the STS-2DTooth dataset consists mainly of anonymized PXI data provided by patients in dental hospitals who have signed consent and waiver forms. This is because, even if the PXI data is processed in a way that conceals personal privacy, the public sharing of PXI data without patient consent violates privacy laws. Therefore, the unlabelled dataset for this study is also limited. Second, the percentage of children's teeth in STS-2DTooth is not high, but the fact that children's teeth are more complex in structure and thus more difficult to segment than adults' teeth should be focused on. However, in addition to the difficulty of obtaining data, the task of labeling children's teeth is also more challenging.

In addition, for the CBCT task, there is still some room for improvement in the prediction scores of the top seven participants. On the one hand, the contestants' algorithms still need to be improved, especially the metal artifacts in CBCT, which will largely affect the accuracy of segmentation, so the solution to the artifacts is an urgent problem for the contestants to focus on. On the other hand, the relatively small number of trainable labelled CBCTs also contributes to the generally low performance of the players. This is due to the inherent difficulty of CBCT annotation; a patient's CBCT data has 400 slices, which requires several annotating doctors to spend a lot of time on annotation.

Finally, none of the AI algorithms presented in this challenge can provide testing for clinical use. We hope that these algorithms will be further evaluated and tested, especially by providing them for use by physicians to compare their predictive effects. Therefore, these AI algorithms still have a long way to go before they are mature enough to be used in software development.

\section{Conclusion}
The challenge, named STS 2023, aims to explore the application of tooth segmentation from two image types. All participants were required to choose between 2D PXI-based or 3D CBCT-based tooth segmentation.
They faced problems such as low-labelled data, which called for the use of semi-supervised algorithms to utilize the large amount of unlabelled data provided.
In both the 2D and 3D tasks, the top-ranked teams used the semi-supervised algorithm system.
Their scores were only affected by their backbone network selection and the specific details of the implementation of the semi-supervised algorithms.
It is worth mentioning that the multi-stage training strategy showed excellent performance in the 3D segmentation task.
However, many teams did not employ any semi-supervised strategies.
Therefore, future STS Challenges may focus on enabling players to adopt semi-supervised algorithms and increase their usage to cope with the existing problems related to the lack of annotation and difficulty in the annotation of dental medical images.

\section*{CRediT authorship contribution statement}
\textbf{Yaqi Wang}: Conceptualization, Investigation, Funding acquisition
\textbf{Yifan Zhang}: Data Curation, Resources
\textbf{Xiaodiao Chen}: Investigation, Methodology
\textbf{Shuai Wang}: Software, Validation
\textbf{Dahong Qian}: Supervision, Project administration
\textbf{Fan Ye}: Writing-Original Draft, Visualization
\textbf{Feng Xu}: Writing-Original Draft, Visualization
\textbf{Hongyuan Zhang}: Writing-Review \& Editing,Visualization
\textbf{Qianni Zhang}: Investigation, Formal analysis
\textbf{Chengyu Wu}: Writing-Review \& Editing, Validation
\textbf{Yunxiang Li}: Investigation, Project administration
\textbf{Weiwei Cui}: Methodology
\textbf{Shan Luo}: Data Curation
\textbf{Chengkai Wang}: Investigation
\textbf{Tianhao Li}: Data Curation
\textbf{Yi Liu}: Funding acquisition
\textbf{Xiang Feng}: Investigation
\textbf{Huiyu Zhou}: Supervision
\textbf{Yundong Liu}: Competitor
\textbf{Qixuan Wang}: Competitor
\textbf{Zhouhao Lin}: Competitor
\textbf{Wei Song}: Competitor
\textbf{Yuanlin Li}: Competitor
\textbf{Bing Wang}: Competitor
\textbf{Chunshi Wang}: Competitor
\textbf{Qiupu Chen}: Competitor
\textbf{Mingqian Li}: Competitor

\section*{Acknowledgements}
This work was supported by the National Natural Science Foundation of China (No.62206242) and China Science and Technology Foundation of Sichuan Province \\ (No.2022YFS0116). There are no conflicts of interest between authors. Yifan Zhang is the principal sponsor of the challenge by collecting and providing clinical data.
Only the organizers and members of their immediate team have access to test case labels.

\bibliographystyle{cas-model2-names}
\bibliography{refs}



\end{document}